\ifx\pdfsuppresswarningpagegroup\undefined\else\pdfsuppresswarningpagegroup=1\relax\fi
\documentclass[fleqn,10pt]{wlscirep}
\usepackage[utf8]{inputenc}
\usepackage[T1]{fontenc}

\usepackage{xr-hyper}
\expandafter\let\csname l@subsubsection\endcsname\undefined
\usepackage{subcaption}
\makeatletter

\usepackage{multicol}
\usepackage{multirow}
\usepackage{graphicx}
\graphicspath{{figs/}}
\usepackage{textcomp}
\usepackage{xcolor}
\usepackage{booktabs}
\usepackage{tikz}
\usepackage{pgfplots}
\RequirePackage{hyperref}
\usepackage[utf8]{inputenc} 
\usepackage{subcaption}
\usepackage{scalerel}
\usepackage{numprint}
\usepackage[capitalize]{cleveref}

\usepackage{relsize}
\usepackage[shortcuts,style=tree,nonumberlist,nopostdot]{glossaries}

\makenoidxglossaries
\setacronymstyle{long-short}
\newcommand\acrodef[2]{\newacronym{#1}{#1}{#2}}

\acrodef{CNN}{convolutional neural network}
\acrodef{ePath}{electronic pathology}
\acrodef{LIME}{local interpretable model-agnostic explanations}
\acrodef{PCA}{principal component analysis}
\acrodef{KDE}{kernel density estimate}
\acrodef{GELiPKeSaM}{global explainability through linear projection of keywords from saliency maps}
\acrodef{ORNL}{Oak Ridge National Laboratory}
\acrodef{LANL}{Los Alamos National Laboratory}
\acrodef{NCI}{US National Cancer Institute}
\acrodef{NIH}{National Institutes of Health}
\acrodef{AJCC}{American Joint Committee on Cancer}
\acrodef{SEER}{Surveillance, Epidemiology, and End Results}
\acrodef{DNN}{deep neural network}
\acrodef{DAC}{deep abstaining classifier}
\acrodef{AI}{artificial intelligence}
\acrodef{JDACS4C}{Joint Design of Advanced Computing Solutions for Cancer}
\acrodef{DOE}{United States Department of Energy}
\acrodef{MTCNN}{multitask convolutional neural network}
\acrodef{SHAP}{SHapley Additive exPlanations}
\acrodef{ML}{machine learning}
\acrodef{DL}{deep learning}
\acrodef{NLP}{natural language processing}
\acrodef{PHI}{personal healthcare information}
\acrodef{PII}{personal identifiable information}
\acrodef{IRB}{institutional review board}
\acrodef{SME}{subject matter expert}
\acrodef{GradInp}{gradient $\bullet$ input}
\acrodef{PC}{principal component}
\acrodef{ALE}{aggregated local explanations}

\newcommand\edit[1]{{\color{black}#1}}

\newcommand\editone[1]{{\color{black}#1}}

\newcommand\secondedit[1]{{\color{black}#1}}

\newif\ifnotes \notestrue

\newcommand*{\addFileDependency}[1]{
\typeout{(#1)}
\@addtofilelist{#1}
\IfFileExists{#1}{}{\typeout{No file #1.}}
}\makeatother

\newcommand*{\myexternaldocument}[1]{%
\externaldocument{#1}%
\addFileDependency{#1.tex}%
\addFileDependency{#1.aux}%
}

\myexternaldocument{SI}
\makeatletter
\let\r@LastPage\undefined
\makeatother

\title{Global explainability of a deep abstaining classifier}

\author[1,2,$\ast$]{Sayera Dhaubhadel}
\author[1]{Jamaludin Mohd-Yusof}
\author[1]{Benjamin H. McMahon}
\author[2]{Trilce Estrada}
\author[1]{Kumkum Ganguly}
\author[3]{Adam Spannaus}
\author[3]{John P. Gounley}
\author [4]{Xiao-Cheng Wu}
\author [5]{Eric B. Durbin}
\author[3]{Heidi A. Hanson}
\author[1,$\ast$]{Tanmoy Bhattacharya}
\affil[1]{Los Alamos National Laboratory, Los Alamos, NM 87545, USA}
\affil[2]{University of New Mexico, Albuquerque, NM 87131, USA}
\affil[3]{Oak Ridge National Laboratory, Oak Ridge, TN 37830, USA}
\affil[4]{Louisiana Tumor Registry, New Orleans, LA 70122, USA}
\affil[5]{Kentucky Cancer Registry, Lexington, KY, 40504, USA}

\affil[*]{sayeradbl@lanl.gov; tanmoy@lanl.gov}

\begin{abstract}
We present a global explainability method to characterize sources of errors in the histology prediction task of our real-world \ac{MTCNN}-based \ac{DAC}, for automated annotation of cancer pathology reports from \acs{NCI}-\acs{SEER} registries. 
Our classifier was trained and evaluated on 1.04 million hand-annotated samples and makes simultaneous predictions of cancer site, subsite, histology, laterality, and behavior for each report. 
The \ac{DAC} framework enables the model to abstain \secondedit{on} ambiguous reports and/or confusing classes to achieve a target accuracy on the retained (non-abstained) samples, but at the cost of decreased coverage.  
Requiring 97\% accuracy on the histology task caused our model to retain only 22\% of all samples, mostly the less ambiguous and common classes.
Local explainability with the \textit{\acs{GradInp}} technique provided \secondedit{a} computationally efficient way of obtaining contextual reasoning for thousands of individual predictions.  
Our method, involving dimensionality reduction of approximately 13000 aggregated local explanations, enabled \secondedit{global} identification of sources of errors  as \textit{hierarchical complexity among classes}, \textit{label noise}, \textit{insufficient information}, and \textit{conflicting evidence}.
This suggests several strategies such as exclusion criteria, focused annotation, and reduced penalties for errors involving hierarchically related classes to iteratively improve our \ac{DAC} in this complex real-world implementation.

\end{abstract}

\pgfplotsset{compat=1.18}

\begin{document}

\flushbottom
\maketitle
\section*{INTRODUCTION}
\label{sec:intro}

\editone{The \ac{DOE} and \secondedit{the} \ac{NCI} have collaborated \cite{nci-doe} to develop models for automated annotation of cancer pathology reports \cite{alawad2019mtcnn, gao18, dhaubhadel2022, peluso24} to assist the \ac{SEER} cancer registries \cite{SEER}. 
In this work, we analyze a model consisting of an \ac{MTCNN} classifier, augmented with an abstention framework.}
This \ac{MTCNN} \ac{DAC} simultaneously makes predictions on five tasks of interest to the \ac{SEER} registries: primary site (70 classes), histological type (599 classes), primary subsite (325 classes), laterality (7 classes), and behavior (4 classes) to achieve an overall accuracy higher than 97\%\edit{, a requirement from the \ac{NCI} \ac{SEER} program} for it to be eligible for use in the real world. While recent advancements in \acs{NLP} research has shown impressive capabilities \cite{openai2024gpt4, touvron2023llama}, achieving 
\edit{this} accuracy with real world observational data, \secondedit{containing} biases, noise, imperfect ground truth, and variation in document structures, still remains challenging \cite{nguyen2015}.

Our desire to improve the performance of our automated classifier has resulted in the development and comparison of multiple model architectures \cite{alawad2019mtcnn, gao18, chandrashekar2024}.  In this work, we aim to improve the performance of our model by investigating and improving different aspects of the workflow for training our classifier using the historically annotated pathology reports.  These aspects range from data staging, cleaning, and labeling to visualization of outputs.  Doing this requires an understanding of the logic by which reports are classified or misclassified.  We call this explainability.

We have found our \ac{DAC} framework \cite{dhaubhadel2022, thulasidasan2019}, described in detail in the Methods section, to be critical in this effort.  Our framework adds an extra class with no training data \secondedit{for} each task by modifying the loss function, so that no answer is provided (model abstention) unless it can statistically guarantee the desired accuracy. 
Obtaining over 97\% accuracy, however, came with a significant cost to coverage (i.e., fraction of non-abstained instances), leading to 100\% abstention for less prevalent or otherwise confusing classes across the classification tasks.
Hence, at the core of this work is a novel workflow that integrates both global and local explainability techniques to understand the reasons for abstention and ultimately enhance the classification of cancer pathology reports. 
By employing dimensionality reduction techniques like \ac{PCA}, the approach simplifies the interpretation of \secondedit{a} large number of local explanations, enabling the identification of meaningful global patterns, such as conflicting information and label noise, paving the way for targeted improvements in automated cancer report annotation systems. 

Our previous work \cite{dhaubhadel2022} examined confusion in cancer sites, such as lung cancer confused with breast cancer, identifying both confusing anatomy and metastatic tumors as sources of classification error.  It used a statistical analysis of \ac{LIME} \cite{ribeiro2016lime} outputs, but noted the computational expense required to compute stable results on the comparatively lengthy pathology reports.  
\edit{In this work, we focus} on histology confusion from breast and lung cancers, which could provide the largest potential \secondedit{coverage improvement}
for our \ac{NCI}-\ac{SEER} information extraction workflow.  
The complexity of cancer histology classification arises in part from the sheer number of histology types and anatomical subsites, documented in detail in the \acs{AJCC} staging manual \cite{ajccV9, ajccV1}.
Additional complexity arises from the steady evolution of the number of classes and their definitions over time as our understanding of cancer grows, leading to lumping, splitting, and \secondedit{introducing}
entirely new classes \cite{amin2017}. 

In this work, we present a global explainability method that will enable \secondedit{the} identification of specific strategies to iteratively improve the performance of a \ac{DL} classifier.
We demonstrate the utility of this method for our \ac{MTCNN} \ac{DAC} classifier that automatically annotates \secondedit{the anatomic} site, histology, subsite, laterality, and behavior of cancer given the pathology reports\edit{, ensuring the target accuracy of at least 97\% for each task}.
We present our results in the context of identification of the four most common histology types of two of the most prevalent cancer types -- lung and breast cancers.  This work identifies mismatches between ground truth and predicted labels for reports due to \edit{complex hierarchical relationships between labels}, label noise, ambiguous/conflicting information, and lack of \edit{relevant} information. 
This framework enables us to identify potential improvements in training protocols, such as \secondedit{excluding}
reports with label noise or insufficient information, and better curate reports with conflicting or ambiguous information.

\section*{RESULTS}
\label{sec:results}

We first compare our baseline \ac{MTCNN} (without abstention) results to our \ac{MTCNN} \ac{DAC}, tuned to achieve 97\% accuracy and note the 78\% abstention rate for histology, motivating our study.  We then characterize the lung and breast cancer histology problems in some detail, showing how the \ac{DAC} greatly simplifies the decision-space by abstaining on the confusing samples and identifying the most important lung and breast cancer histology classes.  Finally, we show how to compute local explainability, then aggregate them into meaningful global insights. \editone{While our observations may generalize more broadly, all subsequent discussions only pertain to lung and breast cancers}.


\begin{table}[htbp]
    \begin{center}
        \caption{
        Accuracy \edit{(micro F1 score)} and abstention rates across five classification tasks for the \ac{MTCNN} \ac{DAC} model tuned to a 97\% target accuracy. The table compares baseline model and \ac{DAC} performance. While the \ac{DAC} achieves the desired accuracy for all tasks, abstention rates vary widely, especially for histology and subsite tasks, reflecting the challenges of complex classifications. 
        }
        \vspace*{-0.5\baselineskip}
        \label{tab:abs_acc}
        \begin{tabular}{|c|c|c|c|c|c|c|c|}
        \toprule
            Model & Tasks  &  Site  &  Histology  &  Subsite  &  Laterality  &  Behavior  &  Overall\\
            \hline
            Baseline & Accuracy  &  91.8  &  78.6  &  68.7  &  91.7  &  98   & 52.2 \\
            \hline
            \ac{DAC} & Accuracy  &  97.6  &  97.45  &  97.48  &  97.59  &  97.87   &  97.46  \\
            & \% abstained  &  17.54  &  78.54  &  72.03  &  21.46  &   0.29  &  87.32  \\
            
        \bottomrule
        \end{tabular}
        \vspace*{-1\baselineskip}
    \end{center}
\end{table}

\cref{tab:abs_acc} shows the abstention and accuracy across the five tasks in our \ac{MTCNN} \ac{DAC} classifier. A baseline \ac{MTCNN} model performs with widely varying accuracy across the five tasks ranging from 69\%-98\%. However, the overall accuracy drops to 52.2\% when considering all five tasks simultaneously.
Using a \ac{MTCNN} \ac{DAC} model enabled slightly over the desired 97\% accuracy on all five tasks for the retained samples. However, tuning the \ac{MTCNN} \ac{DAC} for a target accuracy of 97\% resulted in variable abstention rates across tasks, nearing 80\% for the histology and subsite tasks. 
Moreover, the fraction of retained samples (i.e. coverage) for which the model made predictions on all five tasks simultaneously without abstention was only 13\%.    
\editone{Previous work~\cite{alawad2019mtcnn} has shown a great variability in classification accuracy across tasks depending on the 
complexity of the task}. For the rest of the study, we will examine and present results of histology classification of lung and breast cancers in more detail.

\subsection*{Histology Classification}
Table \ref{tab:HistStats} characterizes the four most common histology classes for lung (137,438 samples) and breast (264,947 samples) cancers, which together account for roughly 40\% of \edit{the} 1.04M samples in the study. The four most common histology classes account for 82\% of all lung cancers and 90\% of all breast cancer reports.

\begin{table}[htbp]
	\begin{center}
		\caption{Performance comparison for the most common lung and breast cancer histology classes under baseline and \ac{DAC} models. The table shows \edit{total number of samples,} precision (PPV), recall (sensitivity), and abstention rates for the presented classes. \edit{Note that each sample has only one ground truth label i.e. the hierarchical relationship is not captured in the labeled data.} 
        For the \ac{DAC}, the precision (PPV) and recall (Sens) are computed for the retained (non-abstained) samples. 
        \edit{NA entries in the PPV columns correspond to classes that were full abstained, and consequently have zero sensitivity.}
        The \ac{DAC} simplifies decision-making by focusing on dominant classes (e.g., `adenocarcinoma’ and `ductal carcinoma’) while abstaining from rarer or more ambiguous classes (e.g., `mixed ductal and lobular carcinoma’).
        Indicated abbreviations are used for clarity in subsequent figures. 
        'No. samples' are the number of reports of the indicated site and histology present in the study.  
        }
        \label{tab:HistStats}
        \footnotesize
	\begin{tabular}{|c|c|l|c|c|c|c|c|c|c|}
        \toprule
                      & Histology   &                  &         &  No.  &  Baseline & Baseline & DAC   &DAC & DAC\\
        \textbf{Site} & Code        &     Histology    & Abbrev. &   samples   & PPV   &  Sens &  PPV & Sens  & Abst  \\
        \hline
         & 8140 & Adenocarcinoma                         &  AdCa   &  54,821 &  76.0 &  93.5  &  97.8 & 12.1 &  87.8  \\
         & 8070 & Squamous Cell Carcinoma                &  SqCa   &  32,012 &  85.4 &  85.0  &  97.3 & 22.7 &  77.1  \\
 Lung    & 8041 & Small Cell Carcinoma                   &  SmCa   &  17,255 &  91.2 &  89.2  &  97.9 & 41.2 &  58.7 \\
         & 8046 & Non-Small Cell Carcinoma               &  nSmCa  &   8,695 &  58.4 &  39.8  &  NA   &  0.0 &  99.2  \\
         & ... &  Other classes                          &   -     &  24,655 &  61.6 & 38.3   &  66.0 &  2.5 &  96.2  \\
       \hline
         & 8500 & Ductal Carcinoma                       &  DuCa   & 193,345 & 86.6 &  95.3 &  98.1  & 39.8 & 60.2 \\
         & 8520 & Lobular Carcinoma                      &  LoCa   &  20,596 & 81.8 &  79.7  &  97.0  & 13.7 & 85.7 \\
 Breast  & 8522 & Mixed Ductal and Lobular Carcinoma     &  MxDuLo &   9,258 & 65.9 &  38.6  &  NA &  0.0 & 94.1  \\
         & 8523 & Intraductal Ca mixed with other types  &   DuOth &  15,323 & 55.0 &  40.0 &  NA  & 0.0 & 98.0   \\
         & ... & Other classes                           &   -     &  26,425 & 57.3   &   34.0 &  4.8   &   0.1  & 97.6  \\
        \bottomrule
        \end{tabular}
        \vspace*{-2\baselineskip}
	\end{center}
\end{table}

The effect of abstention on the lung and breast cancers studied in this work is shown in Table \ref{tab:HistStats}  
\editone{, with the largest classes named and} the remaining minority classes lumped together as \textit{other classes}. 
A baseline (non-abstaining) \ac{MTCNN} classifier is able to achieve 76\% accuracy on lung cancer adenocarcinoma and 87\% accuracy on breast cancer ductal carcinoma, the most prevalent histology classes in lung and breast cancers respectively. 
We see in the histology confusion matrices in Supplementary Figures \ref{fig:confusion_matrix} (left panel) that at least 20 different lung and breast histology classes are mislabeled as adenocarcinoma and ductal carcinoma, respectively, \editone{suggesting that less common classes are often misclassified as a more common histologic subtype}. On using an \ac{MTCNN} \ac{DAC} tuned for 97\% accuracy, the accuracy goes over 97\% for non-abstained reports, but 88\% of the lung adenocarcinomas and 60\% of the breast ductal carcinomas are abstained.  

\edit{The \ac{MTCNN} \ac{DAC} model does not predict on the minority classes, with only the three most common lung and two most common breast histologies having a an abstention rate below 90\%}.
For both \textit{non-small cell carcinoma} and \textit{mixed ductal and lobular carcinoma}, requiring a 97\% accuracy prevents the \ac{DAC} from classifying \textbf{any} reports belonging to these classes. This illustrates the \ac{DAC}'s tendency to simplify the decision surface 
\edit{by abstaining on confusing/ambiguous samples}.

\editone{Table \ref{tab:HistStats} also shows two types of complexities in the histology classification for these sites. First, histology classes can have complex and overlapping names, for example, \textit{`ductal carcinoma'} and \textit{`mixed ductal and lobular carcinoma'}. Second, there may be hierarchical relationship\secondedit{s} between classes}.  Consideration of the \ac{SEER} detailed guidance for lung~\cite{seerLung} and breast~\cite{seerBreast} cancers provides some insight into the potential difficulties.  
For example, adenocarcinoma (code \textit{8140}) and squamous cell carcinoma (code \textit{8070}) are specific types of non-small cell carcinoma (code \textit{8046}) and the coding guidance~\cite{seerLung} states \textit{``code the specific histology when the diagnosis is non-small cell lung carcinoma (NSCLC) consistent with (or any other ambiguous term) a specific carcinoma (such as adenocarcinoma, squamous cell carcinoma, etc.) when: i) the histology is clinically confirmed by a physician (attending, pathologist, oncologist, pulmonologist, etc.), ii) the patient is treated for the histology described by an ambiguous term. Note 1: If the case does not meet the criteria in the first two bullets, code non-small cell lung cancer (NSCLC) 8046"}. This guidance clearly highlights the need for information outside the pathology report \editone{in case of ambiguous language,} to accurately ascertain the histology class.

\editone{An additional source of confusion may arise because a \ac{SEER} registrar may have determined the ground truth based on information from multiple sources and/or a combination of multiple pathology reports, while the \ac{DAC} is learning from information only in the pathology reports, and making predictions on a single report at a time.}

\subsection*{Local explanations with \textit{\ac{GradInp}} technique}

\begin{figure}[!htb]
    \vspace*{-0.5\baselineskip}
    \captionsetup[subfigure]{justification=centering, font=small}
    \centering
        \begin{subfigure}{\textwidth}
        \vspace*{-1\baselineskip}
        \caption{}
	\vstretch{1}{\includegraphics[width=\columnwidth]{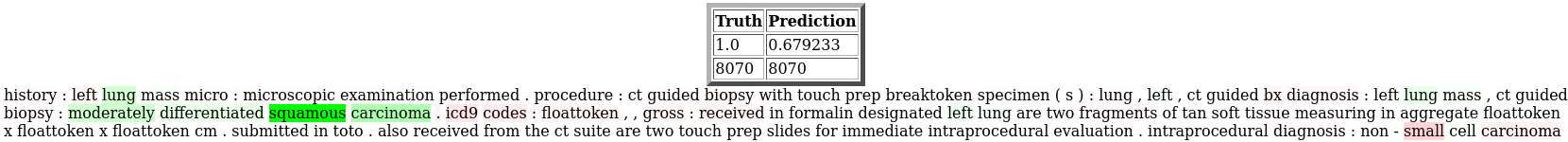}}
            \label{fig:lu_report}
        \end{subfigure}
        \begin{subfigure}{\textwidth}
        \vspace*{-1\baselineskip}
	\caption{}
	\vstretch{1}{\includegraphics[width=\columnwidth]{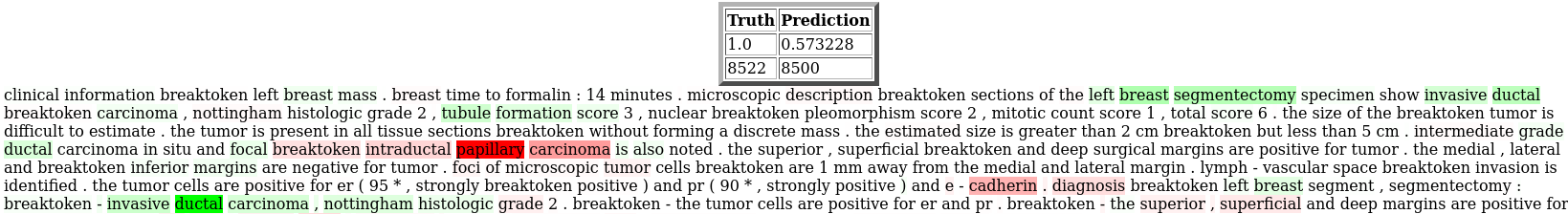}}
            \label{fig:br_report}
        \end{subfigure}
        \begin{subfigure}{.245\textwidth}
		 \vstretch{1.3}{\includegraphics[width=\columnwidth]{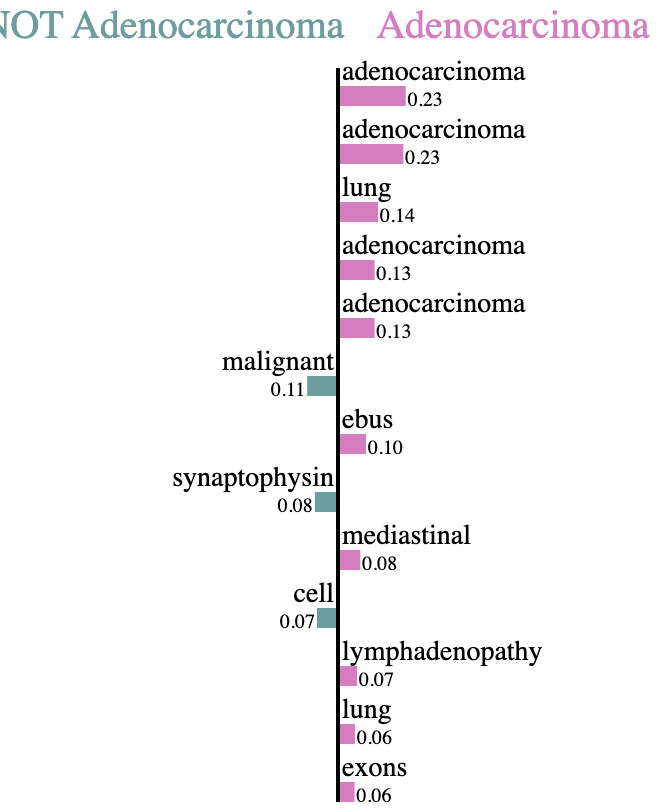}}
		\caption{AdCa predicted as AdCa}
            \label{fig:AdcAd}
	\end{subfigure}
        \begin{subfigure}{.25\textwidth}
	     \vstretch{1.2}{\includegraphics[width=\columnwidth]{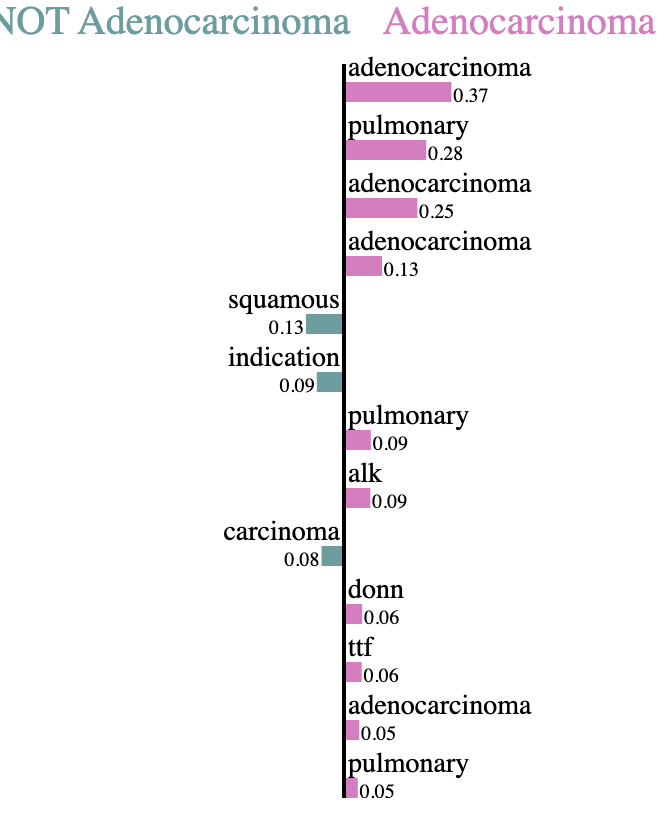}}
            \caption{AdCa predicted as AdCa with \newline conflicting information (\textit{Squamous})}
		\label{fig:AdcAd_confusing}
        \end{subfigure}
        \begin{subfigure}{.245\textwidth}
		 \vstretch{1.065}{\includegraphics[width=\columnwidth]{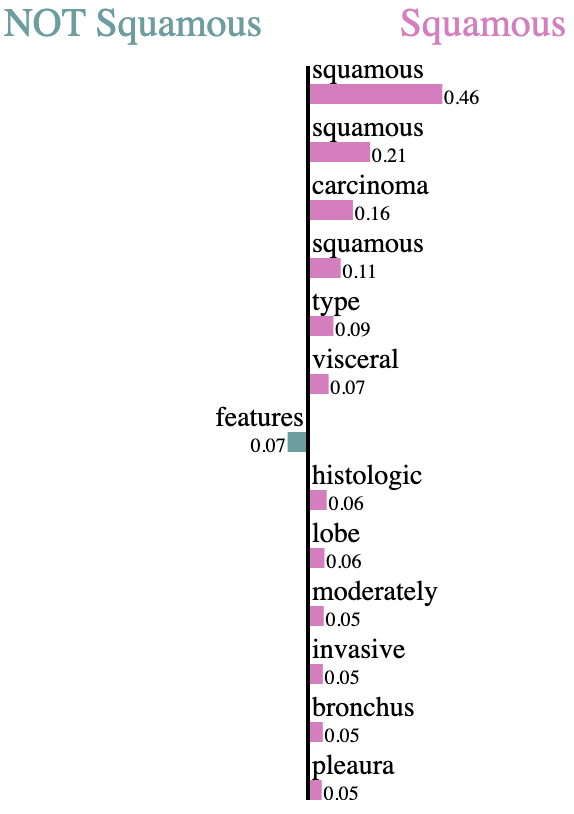}}
         \caption{AdCa predicted as \\ Sq cell Ca (label noise)}
		\label{fig:AdcSq}
	\end{subfigure}
        \begin{subfigure}{.245\textwidth}
	     \vstretch{0.99}{\includegraphics[width=\columnwidth]{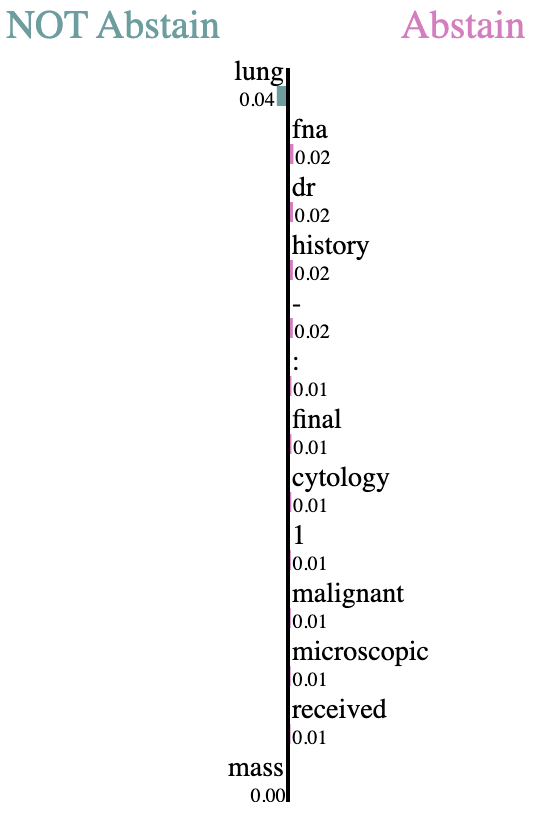}}
            \caption{AdCa abstained - no histology information}
		\label{fig:AdcAbs}
        \end{subfigure}
    \caption{
    Local explanations of the model’s classification decisions using the \textit{\acs{GradInp}} technique. Panels (a) and (b) highlight key words in sample pathology reports that support or oppose the model’s predictions, \editone{indicated by the sign of the weight}. Panel (a) shows a correctly classified squamous cell carcinoma (\textit{8070}) in the presence of hierarchical complexity (e.g., `non-small cell carcinoma’), while panel (b) illustrates a mixed ductal and lobular carcinoma (\textit{8522}) incorrectly classified as ductal carcinoma \textit({8500}), demonstrating label noise. Panels (c)–(f) present bar charts of the top 13 (out of 20) words influencing predictions for lung cancer adenocarcinoma cases, including correct and incorrect classifications. Rightward (magenta) bars indicate evidence supporting the predicted class \editone{(positive \textit{\ac{GradInp}} weight)}, while leftward (olive) bars show evidence opposing it \editone{(negative \textit{\ac{GradInp}} weight)}. The importance of a word varies depending on its context within the report. Note the clustering of reports by histology type and the influence of conflicting evidence in misclassified and abstained cases.
    }
    \label{fig:local_exp}
    \vspace*{-\baselineskip}
\end{figure}

In \cref{fig:local_exp}, we present local explainability results computed using the \textit{\ac{GradInp}}~\cite{shrikumar2017, denil2014} for our \ac{MTCNN} \ac{DAC} model tuned to achieve 97\% accuracy in two formats -- i) the original report with its text highlighted based on the local explainability weights, and ii) explainability weights in signed and sorted bar chart format.  
\cref{fig:lu_report} and \cref{fig:br_report} show a \edit{correctly classified} lung cancer squamous cell carcinoma report and an \edit{incorrectly classified (not abstained)} breast cancer mixed ductal and lobular carcinoma report with the words highlighted based on the \textit{\acs{GradInp}} weights. 
\cref{fig:lu_report} presents an example of complexity when classes with hierarchical relationships are treated as independent during training: \textit{squamous cell carcinoma (code 8070)} \edit{is} a more specific type of \textit{non-small cell carcinoma (code 8046)}\edit{, however, the two classes have been formulated as independent during training with the ground truth label for this particular example as \textit{8070}}.
Similarly, \cref{fig:br_report} presents an example of `label noise' as information in the report does not align with the coded ground truth, potentially because it was discerned from other information sources.
These \edit{examples} allow the end users to see the `important features (words)' along with the context in which they are used, enabling evaluation of the model's classification decision at the sample level.
\cref{fig:AdcAd,fig:AdcAd_confusing,fig:AdcSq,fig:AdcAbs} present results in signed bar chart format for four lung cancer adenocarcinoma samples showing the top 20 words with the largest magnitude of explainability weights in the report. These do not explicitly display the context but do provide different weights (sign and magnitude) for multiple instances of the same word in a report depending on the context and are easier to review in larger numbers.

With `manual review' of the local explainability results (in both formats) for a representative sampling pathology reports, we were able to qualitatively assess the model's reasoning for classification. 
While such manual review is appropriate for validating well-defined hypotheses on a pre-selected sampling of reports, it is prohibitively difficult to apply to unsorted or large numbers (thousands) of samples and generate meaningful quantitative insights.



\subsection*{Global explainability}

In this section, we investigate global behavior of our model by aggregating local explanations generated with the \textit{\acs{GradInp}} technique. As shown in \cref{fig:local_exp}, the \textit{\acs{GradInp}} technique assigns distinct weights to each instance of a word in a given report.  If we \editone{aggregate} the weights for all instances of each word (e.g. `adenocarcinoma') in a report, we can create a matrix of scores for the set of most-important words across an illustrative set of pathology reports\editone{, the \ac{ALE} matrix}.  Analysis of this \editone{\ac{ALE} matrix} provides insight into the nature of errors made by our \ac{MTCNN} \ac{DAC}. In constructing this \editone{\ac{ALE} matrix}, we must choose which representative reports and words to include, and how to visualize the information contained within.  This process benefits from the simplification of the decision surface created by use of the \ac{DAC} together with our \ac{MTCNN}, evident by comparing the confusion matrices in Figure \ref{fig:confusion_matrix}.  Each element of the confusion matrix represents a boundary on the histology decision surface.  Use of the \ac{DAC} replaces nearly all of mismatch classes with a single abstention class.

\editone{Below,} we show results for reports with sites correctly classified as lung and breast cancer, separated by their histology prediction, for both non-abstained and abstained samples (using the second highest softmax score after abstention). 
To \edit{minimize the effect of class imbalance in our global explainability results}, we cap the number of samples to 1000 for a given ground truth - prediction combination. Further details are in the Methods section.
Figures \ref{fig:lu_dist} and \ref{fig:br_dist} visualize these data sets, together with $\sim$13000 
reports for each cancer type, plotted and color-coded according to the most important words for each report.  More details are provided in the captions. 

Finally, we applied \ac{PCA} to each \editone{\ac{ALE} matrix} \editone{and investigated the \acp{PC}, cumulatively accounting for 90\% of the variance.} \editone{We observed that the majority of the signal is captured by the first two \acp{PC}}. 
\cref{fig:PCA_Lung,fig:PCA_Breast} show \edit{the first two \acp{PC}} for lung and breast cancers respectively, separated by abstention category (left vs. right panels) and color-coded by the ground truth - prediction combination or keyword presence (top vs. bottom panels). 
\edit{For completeness,} the eigenvectors, eigenvalues, and variance explained by each principal component of the \ac{PCA} computations are provided in Tables \ref{tab:lu_ev_table}, \ref{tab:br_ev_table} for the first five principal components, sorted in descending order of the magnitude along \edit{first two \ac{PC} axes}. We discuss the insights gained from these analyses below, first for lung and then breast cancer.

\subsubsection*{Lung cancer}
\begin{figure}[!htbp]
    \centering
        \begin{subfigure}{0.48\textwidth}
		\includegraphics[width=0.99\columnwidth]{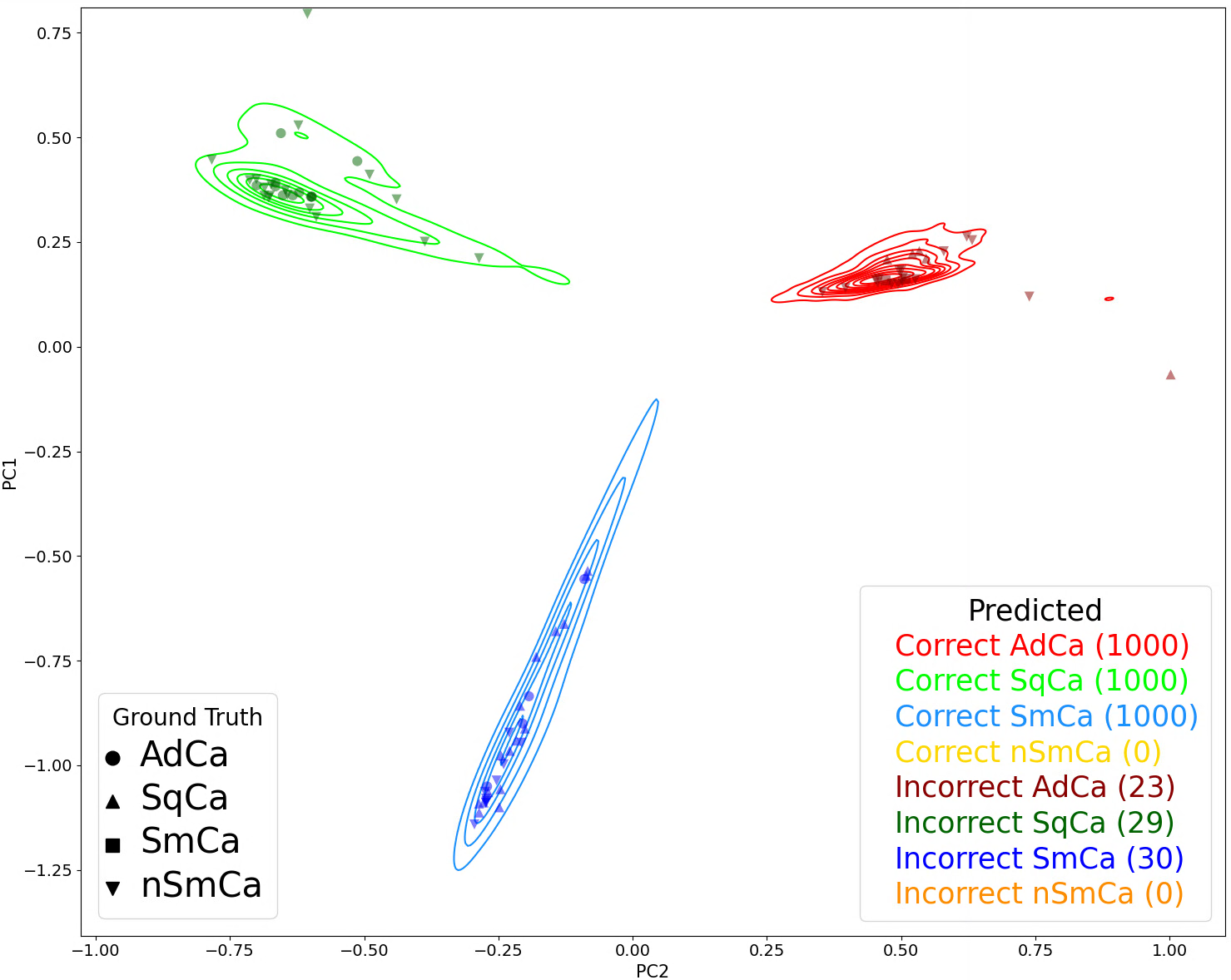}
		\vspace*{-0.5\baselineskip}
            \caption{Non-abstained reports - color coded by their top prediction}
            \label{fig:lu}
	\end{subfigure}
        \begin{subfigure}{0.48\textwidth}
		\includegraphics[width=0.99\columnwidth]{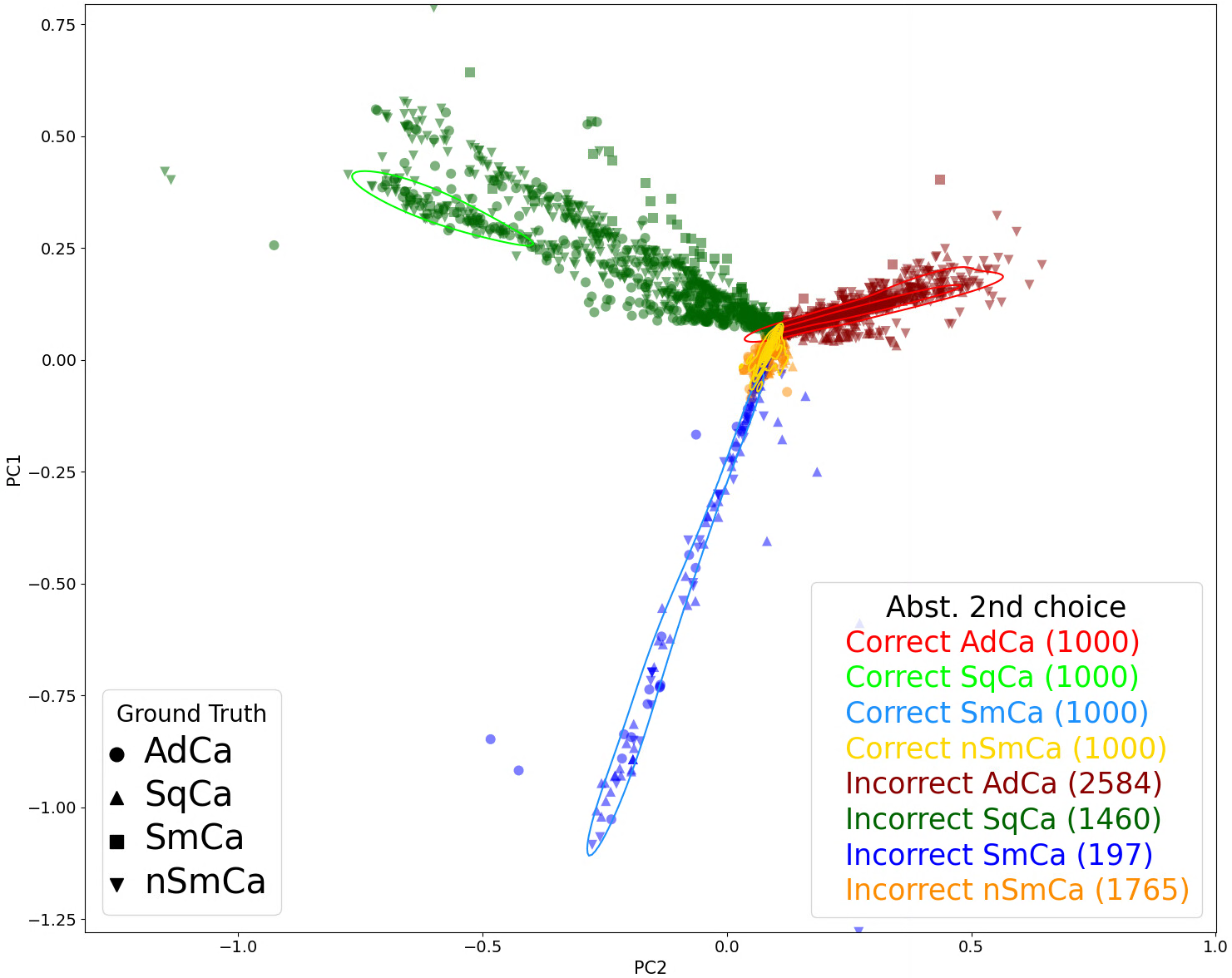}
            \vspace*{-0.5\baselineskip}
            \caption{Abstained reports - color coded by 2nd choice prediction}
		\label{fig:lu_abs}
	\end{subfigure}
        
        \begin{subfigure}{0.48\textwidth}
        \includegraphics[width=0.99\columnwidth]{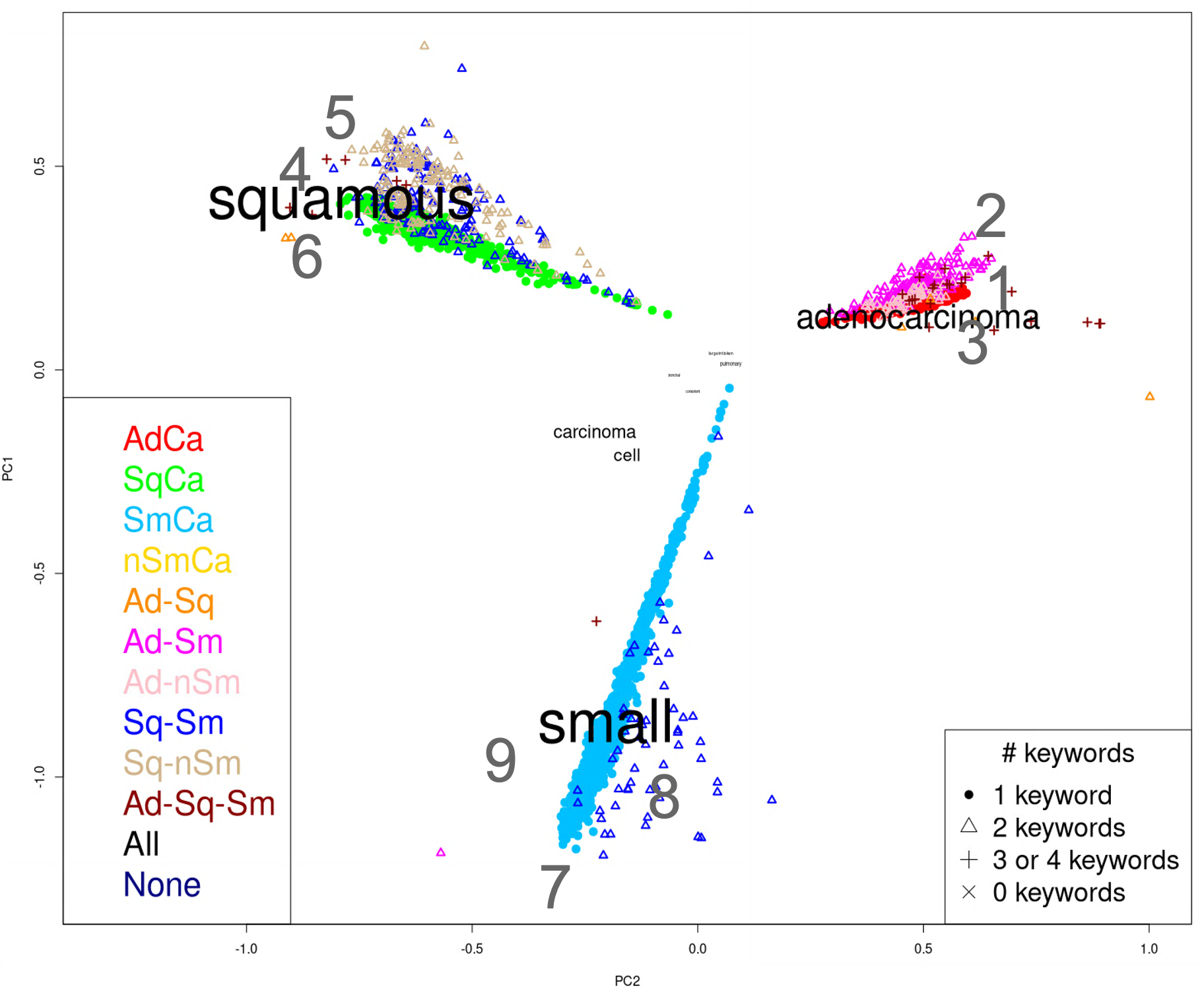}
            \vspace*{-0.5\baselineskip}
            \caption{Non-abstained reports - color coded by class-specific keywords} 
		\label{fig:lu_words}
        \end{subfigure}
        \begin{subfigure}{0.48\textwidth}
        \includegraphics[width=0.99\columnwidth]{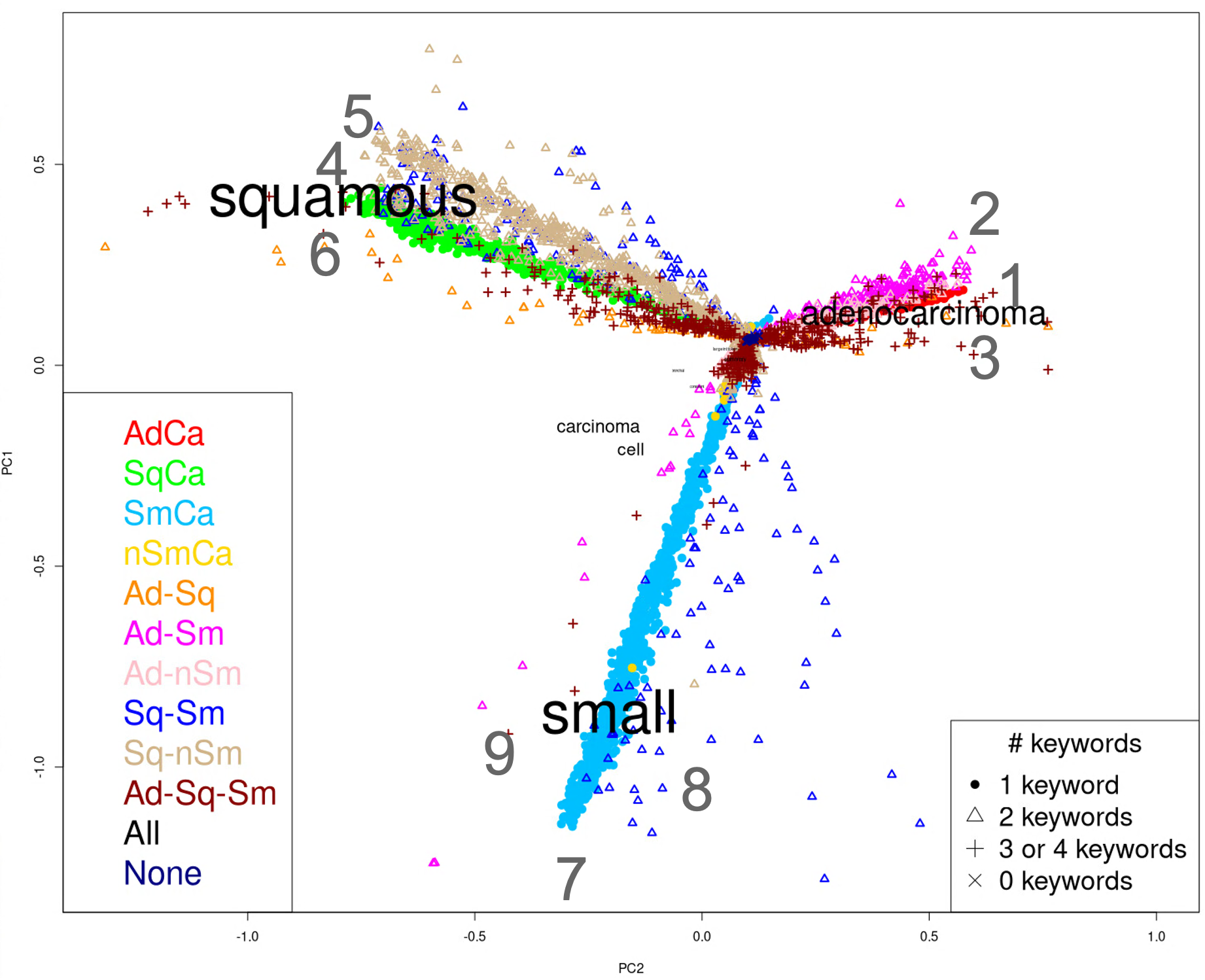}
            \vspace*{-0.5\baselineskip}
            \caption{Abstained reports - color coded by class-specific keywords}
		\label{fig:lu_abs_words}
        \end{subfigure}
    \caption{\edit{First two principal components} from \ac{PCA} analysis of local explanations for the four largest histology classes of lung cancer for \ac{DAC} models trained with 97\% target accuracy for (left) non-abstained and (right) abstained reports showing the first non-abstained choice.  \textbf{(top)} correct classifications are shown with \acrfull{KDE} contours and incorrect classifications with points; color-coded by the predicted class, with lighter hue dots representing correct prediction and darker hue representing incorrect prediction; marker shapes other than dots represent ground truth classes for incorrect predictions. \textbf{(bottom)} all samples shows as points; symbols are color-coded by the presence of the class-specific deterministic keywords obtained from PCA analysis \textit{adenocarcinoma}, \textit{squamous}, \textit{small}, and \textit{non small}; marker shapes represent the number of distinct classes referred to in the report.  Numbers indicate different regions showing specific type of conflicting information. }
    \label{fig:PCA_Lung}
\end{figure}

\cref{fig:PCA_Lung} shows \edit{the plots of the first two \acp{PC}} for the four largest histology classes of lung cancer, with left panels showing non-abstained reports and right panels showing abstained ones; the top and bottom plots show the same data but with a different color scheme.  Each point in the plots represents an individual cancer pathology report. The top panels, Figures \ref{fig:lu}, \ref{fig:lu_abs}, show \ac{KDE} contours for correctly classified reports and scatter points for incorrect ones. They are color-coded based on the predicted class (for non-abstentions) or by their 2nd choice (for abstained samples) -- red hues for adenocarcinoma, green hues for squamous cell carcinoma, blue hues for small cell carcinoma in lung cancer, and yellow hues for non-small cell carcinoma.  
The bottom panels show all reports as points and are color coded according to the presence of the class-specific deterministic keywords identified as the most important by the \ac{PCA} analysis -- \emph{`adenocarcinoma'}, \emph{`squamous'}, \emph{`small'}, and \emph{`nonsmall'} or both \emph{`non'} and \emph{`small'}, with the keywords plotted at the location of their eigenvectors along the \edit{first two \ac{PC}} axes.  

Each point (report) in Fig. \ref{fig:lu} (upper left) in this plot falls in a ray extending from the center, 
corresponding to adenocarcinoma (red), squamous cell carcinoma (green) and small cell carcinoma (blue).  The reports near the center of the \ac{PCA} plot have less total evidence\edit{, as extracted by the \textit{\ac{GradInp}} method,} supporting the classification; the lack of reports near the center of the \ac{PCA} plot indicates the threshold level of evidence required for 97\% confidence in the classification.  Because the individual words defining the cancer histology class provide the majority of evidence for classification  \editone{(Tables \ref{tab:lu_ev_table}, \ref{tab:br_ev_table})}, further insight can be obtained in Figure \ref{fig:lu_words} (bottom left), where reports are color coded according to the presence of one or more of the class-specific keywords.  Narrow rays appear (red, green, skyblue colors in regions 1, 4, and 7) in each of the three regions, corresponding to reports with no conflicting evidence. \edit{The magenta and tan points in regions 1 and 4, overlapping with the red and green points, \editone{appear to be} examples of ambiguity deriving from the hierarchical relationship between classes.}

Further insight into the batches of reports next to these narrow rays (identified in the Fig. \ref{fig:lu_words} as regions 2, 3, 5, 6, 8 and 9) is provided by the observation that these regions all have keywords specific to two or more classes.  For example, region 5 contains both - the word `squamous', associated with the correct classification, as well as `non' and `small' or `nonsmall' or `small', associated with an alternative classification.  Note also that region 5 is shifted up from the reports containing only `squamous', consistent with `small' or `non' and `small' appearing with negative weight in the local explainability calculation for those reports (the positive direction for small is that of the blue ray pointing downward).  A similar pattern is seen for region 2 (`adenocarcinoma' with smaller negative weight for `small' or `non' and `small'), region 3 (`adenocarcinoma' with smaller, negative, weight for `squamous'), region 8 (`small' with smaller, negative, weight for `squamous'), and region 9 (`small' with smaller, negative, weight for `adenocarcinoma'). 

In \cref{fig:lu}, correctly classified reports are shown in lighter colors (e.g. light green for squamous cell carcinoma) while the incorrectly classified reports are plotted on top in a darker color.  Combining this with the reasoning from Fig. \ref{fig:lu_words} described above, two classes of misclassified squamous cell carcinomas are evident.  Those exactly on top of the most dense \ac{KDE} region or region 4 (in \ref{fig:lu_words}) contain information only for one (predicted) class, and thus we refer to them as `label noise', although their precise origin may come from several potential effects.  In contrast, we refer to region 5, above, as containing conflicting information, because of the additional presence of the word `small' or `non' and `small' or `nonsmall' in the reports.  In fact (data not shown), manual analysis of all 29 reports incorrectly predicted as squamous cell carcinoma and plotted in light green in \ref{fig:lu} supports this interpretation.  

Examination of the plots for abstained samples (Fig. \ref{fig:lu_abs} and \ref{fig:lu_abs_words}) shows three types of added complexity.  First, many more incorrectly classified reports (darker hues), second, the addition of a fourth histology class, non-small cell carcinomas (yellow color in \ref{fig:lu_abs}), and third, the presence of all eight types of 2-classes conflicting information (Fig. \ref{fig:lu_abs_words}).
Another qualitative change is the appearance of the `non-small cell' class appearing as offshoots of the `small cell' carcinomas and in between adenocarcinomas and squamous cell carcinomas, rather than radiating from the center like the other classes.  This is \editone{potentially related to} two important factors: i) hierarchical \edit{relationship between classes that are treated as independent i.e.} adenocarcinoma (\textit{8140}) and squamous cell carcinoma (\textit{8070}) two specific types of non-small cell carcinomas (\textit{8046}) \cite{seerLung}, ii) constraint of \ac{PCA} decomposition when there is overlap (the word `small') in `small cell' and `non-small cell' carcinoma. 
Another corollary of this, is that regions 2, 3, 5, 6, 8 and 9 each have separate types of conflicting information, one each for conflicting information for small cell and non-small cell for squamous cell carcinomas and adenocarcinomas.

As we extend our analysis to the abstained reports in Fig. \ref{fig:lu_abs} and look at the second choice classification, i.e the model's potential answer in the absence of the abstention class, we see reports with more complex conflicting information (indicated by brown dots in \cref{fig:lu_abs_words}).  
The categories of conflicting information identified by the model described above for the plots of non-abstained samples, while still visible in plots for the abstained plots, are much less distinct and have more noise.  This is happening both because of the greater complexity of what is decomposed (4 categories instead of three) and the more complex reports that are now under consideration.  Both of these effects argue for the utility of the abstention class in facilitating model interpretability.


\subsubsection*{Breast cancer}
\begin{figure}[!htbp]
    \vspace*{-0.5\baselineskip}
    \centering
        \begin{subfigure}{0.48\textwidth}
            \vspace*{-0.9\baselineskip}
		\includegraphics[width=0.99\columnwidth]{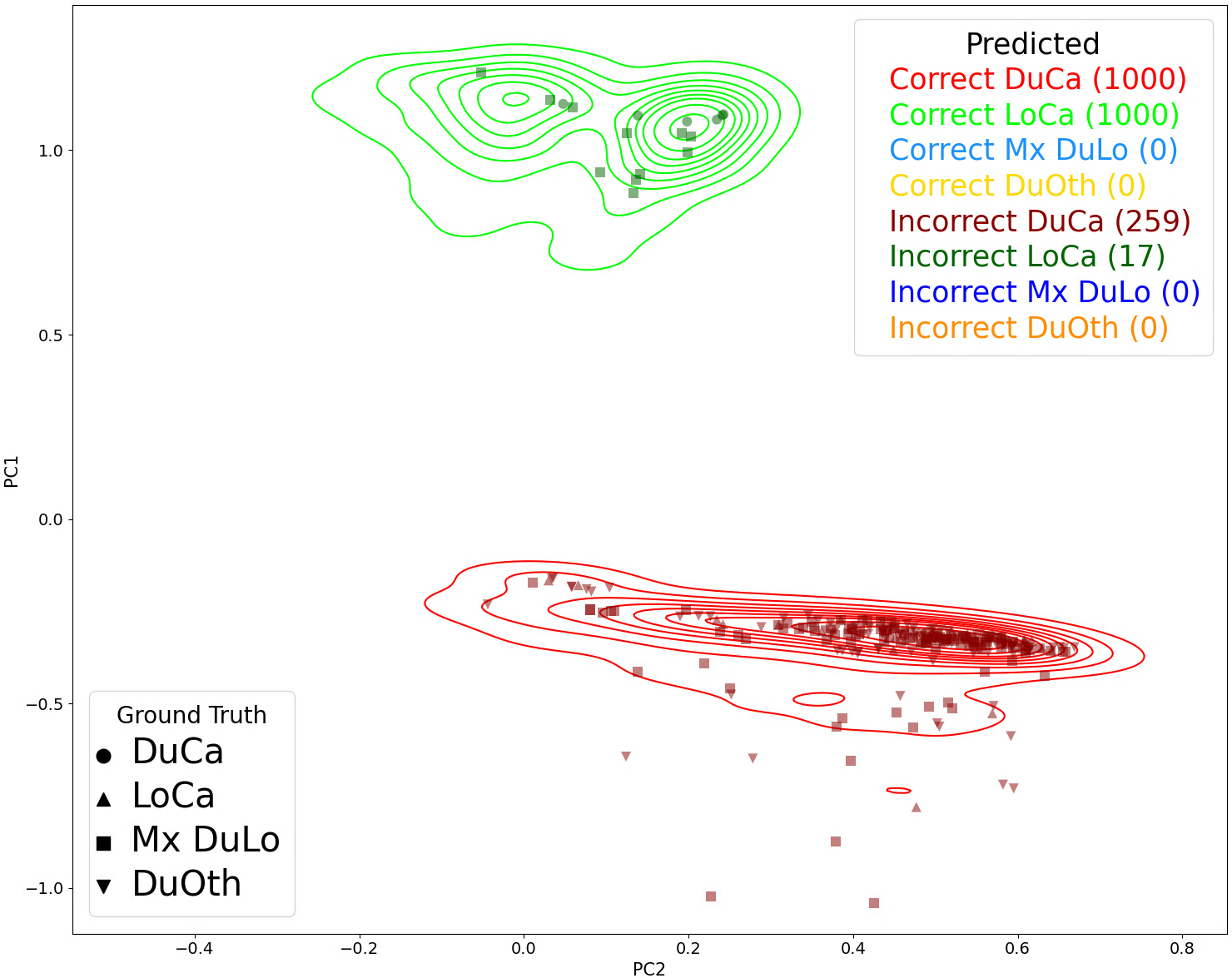}
            \vspace*{-0.5\baselineskip}
            \caption{Non-abstained reports - color coded by their top prediction}
		\label{fig:br}
	\end{subfigure}
        \begin{subfigure}{0.48\textwidth}
            \vspace*{-0.9\baselineskip}
		\includegraphics[width=0.99\columnwidth]{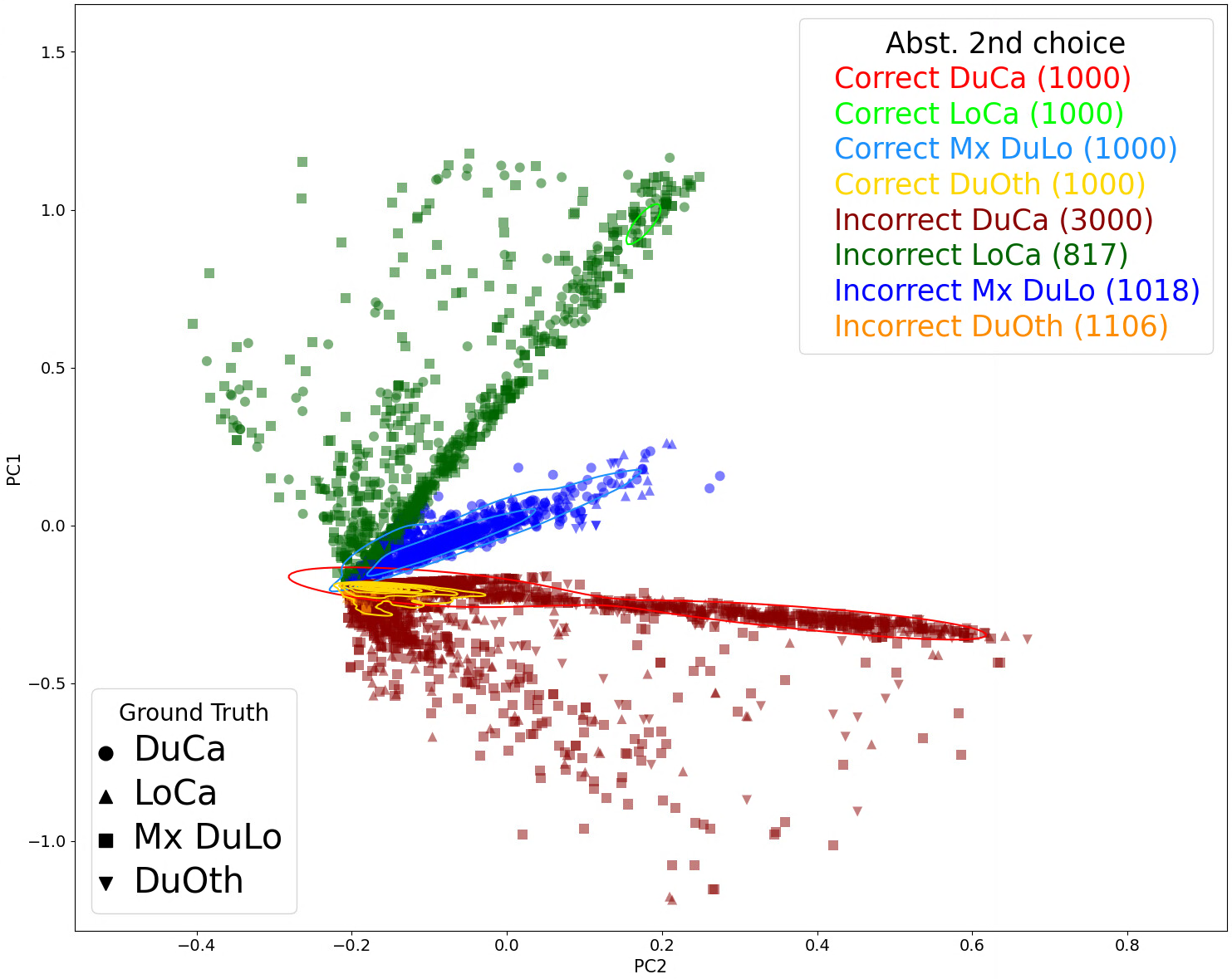}
            \vspace*{-0.5\baselineskip}
            \caption{Abstained reports - color coded by 2nd choice prediction}
		\label{fig:br_abs}
	\end{subfigure}
        
        \begin{subfigure}{0.48\textwidth}
        \includegraphics[width=0.99\columnwidth]{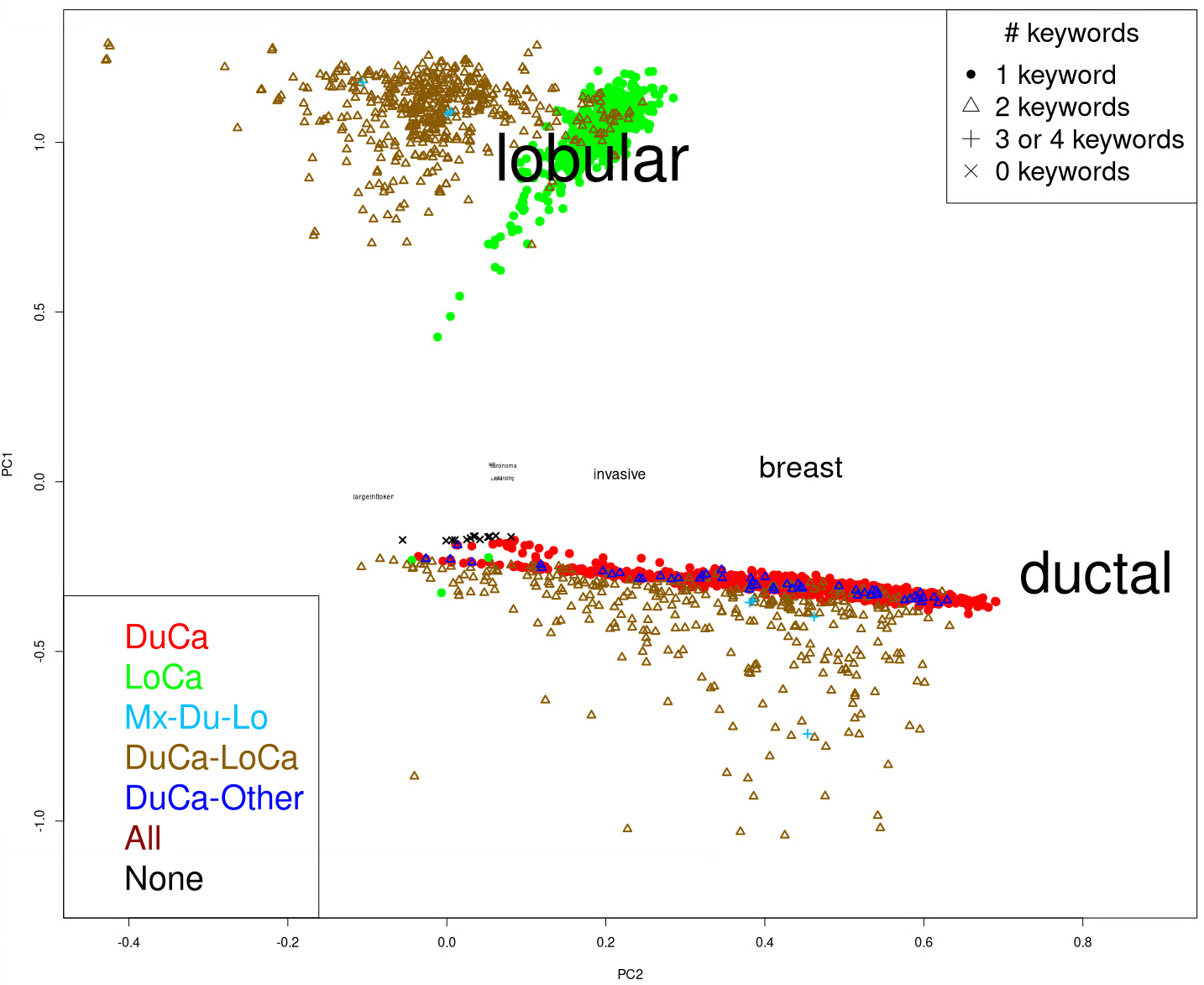}
            \vspace*{-0.5\baselineskip}
            \caption{Non-abstained reports - color coded by class-specific keywords} 
		\label{fig:br_words}
        \end{subfigure}
        \begin{subfigure}{0.48\textwidth}
        \includegraphics[width=0.99\columnwidth]{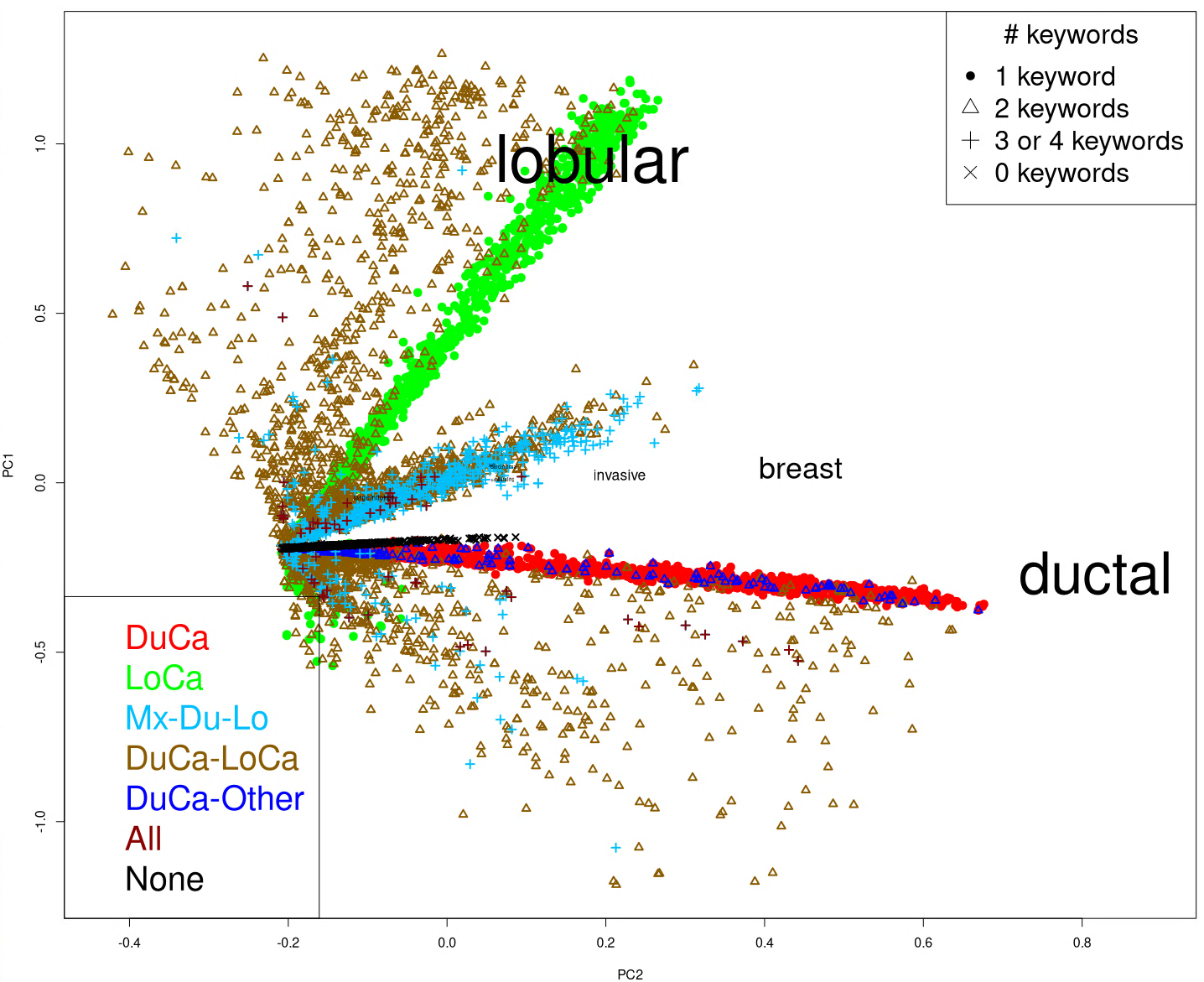}
            \vspace*{-0.5\baselineskip}
            \caption{Abstained reports - color coded by class-specific keywords}
		\label{fig:br_abs_words}
        \end{subfigure}

    \vspace*{-.5\baselineskip}
    \caption{ \textbf{(top)} \edit{First two principal components} from \ac{PCA} analysis of local explanations for the four largest histology classes of breast cancer for \ac{DAC} models trained with 97\% target accuracy for (left) non-abstained and (right) abstained reports showing the first non-abstained choice. \textbf{(top)} correct classifications are shown with \acrfull{KDE} and incorrect classifications with points; color-coded by the predicted class, with lighter hue dots representing correct prediction and darker hue representing mismatching prediction; marker shapes other than dots represent ground truth classes for incorrect predictions.   \textbf{(bottom)} all samples shows as points; color-coded by the presence of the class-specific deterministic keywords \textit{ductal}, \textit{lobular}, \textit{mixed ductal lobular}, and \textit{infiltrating}; marker shapes represent the number of classes referred in the report.  
    The prevalence of ductal carcinoma in the dataset allows the model to correctly predict many cases of ductal carcinoma even in the absence of any histology specific information (black points), based on site information alone.
    }
    \label{fig:PCA_Breast}
\end{figure}

\cref{fig:PCA_Breast} shows the global explainability results for breast cancer.  Plots are constructed as in \cref{fig:PCA_Lung}, but for breast cancer reports.  The red hues indicate ductal carcinoma, green hues indicate lobular carcinoma, blue hues indicate mixed ductal and lobular carcinoma, and yellow hues indicate infiltrating ductal mixed with other carcinomas. The bottom panels are color coded according to the presence of the class-specific deterministic keywords: `ductal', `lobular',  `mixed', or each of the combinations of these words.  

The overall structure of the \ac{PCA} plots for breast cancer shown in \cref{fig:PCA_Breast} is different from the lung cancers, \editone{potentially because of the number of classes under consideration and the relationship between classes}.  
For breast cancer, only two categories, lobular and ductal carcinoma, contribute significantly, and ductal carcinoma comprises over 90\% of these.   Consequently, only two groups of reports radiate out on the \ac{PCA} plot, and not symmetrically.  Further difficulty for our classifier arises from the names of the third and fourth most-prevalent histology classes both being mixtures of ductal carcinoma (see Table \ref{tab:HistStats}). \editone{Accordingly, the third most-prevalent class \textit{`8522 - mixed ductal and lobular carcinoma'} (blue points in \cref{fig:br_abs}) appear in between the `ductal' (red) and `lobular' (green) points, rather than on the other side of the plot.}

Another interesting distinction from lung cancer plots are the reports without any of the three class-specific words (towards the center, indicated by black dots in \cref{fig:br_words} and \cref{fig:br_abs_words}) which are correctly classified as ductal carcinoma, highlighting the prevalence/dominance of ductal carcinoma in the dataset. 
Additionally, tightly focused rays indicate reports with large explainability weights for the indicated class, with errors constituting label noise, or no indication of why the conflicting ground truth exists.  Finally, reports with conflicting evidence can be found next to these rays, with much more diversity for the abstained reports.  The third category - \textit{mixed lobular and ductal carcinoma (8522}) appropriately appears as a ray between those for lobular and ductal, while category four, ductal mixed with other (\textit{8523}) appears at the base of ductal carcinomas.  We see from Table 2 that our baseline (non-abstaining) classifier had a baseline sensitivity of 40\% and PPV of 55-65\% for these classes, while the \ac{DAC} did not attempt to classify them.

\section*{DISCUSSION}
We present a global explainability method designed to identify and characterize dominant sources of errors in a \ac{DL} classifier, demonstrating its use for histology classification in lung and breast cancers in order to improve the performance of our overall workflow.

The \ac{MTCNN} classifier with abstention (\ac{DAC}) \cite{alawad2019mtcnn, dhaubhadel2022} is illustrated in \cref{fig:model_arch}.  It was designed to predict across the full breadth of cancer types and attributes, including 599 histology classes and 325 subsites.  Abstention was originally included as a way to ensure the required 97\% accuracy, albeit with a significant cost in reduced coverage \cite{thulasidasan2019}.  As we set about improving the coverage of our classifier, examination of confusion matrices was of limited value, because of label noise, a complex ground truth with hierarchical relationships between classes, and the large number of categories to investigate, each with individualized complexity.  
This motivated the present work, which both identifies reasons for problematic misclassifications and, through the \ac{DAC}, focuses the model on the most prevalent and problematic histology categories.

Our earlier work \cite{dhaubhadel2022} utilized \ac{LIME} \cite{ribeiro2016lime} to generate local (sample-wise) explanations and used a statistical analysis to identify and validate important concepts for non-abstained as well as abstained samples from a few hundred samples, but was very slow to compute. 
Other local explainability techniques - anchors \cite{ribeiro2018anchors} and \acs{SHAP} \cite{lundberg2017} were also prohibitively slow.
Existing global explainability techniques \cite{lundberg2020,ribeiro2016lime} are based on aggregation of local explanations from \ac{LIME} and \acs{SHAP} respectively and are typically applied on local explanations for a small number of representative samples and essentially are not broadly `global'.
Other methods for topological interpretability \cite{spannaus2024} are in development.
Consequently, we explored gradient based methods such as saliency maps \cite{simonyan2014deepinside, li2016saliencyNLP}, noting the caveats described in \cite{adebayo2018saliencymapsproblem}. We found the \textit{\ac{GradInp}} method \cite{shrikumar2017, denil2014} provided plausible results across a broad range of our report-types with a compute time of 1-2 seconds per sample.  This compute time enabled application across approximately 13,000 reports for this analysis and would also enable interactive inclusion in a real-time report annotation workflow.

While the keywords we identified using our method (such as adenocarcinoma or ductal carcinoma) were strongly correlated with each class and enabled visualizations to gain insight into what the model was doing, it is important to remember that the overall model prediction involves complex non-linear decision boundaries. 
The identification of \edit{hierarchical relationships between classes}, label noise, insufficient information, and conflicting evidence are \edit{enabled by the improved training with \ac{DAC}} as it abstains on the ambiguous/conflicting samples \edit{leading to a cleaner decision boundary}.
Examination of the dominant eigenvectors from our \ac{PCA} will always suggest a list of keywords to aid in the interpretation.
Although developed for our specific, real-world decision support tool, we believe our use case and workflow has several aspects that will generalize to other problems.

\edit{A potential direction for improving the performance may be excluding reports with label noise and insufficient evidence from training sets while better curating reports with conflicting evidence. Likewise, reformulating the classification problem such that predictions within a hierarchy are only partially penalized compared to completely incorrect prediction might help in better training}.

Our study suffers from several limitations.  
First and foremost, we have not actually demonstrated that applications of our suggested workflow lead to substantive improvements in model training.  Such improvements \edit{may be possible, although with considerable additional work as outlined above, and will be the subject of future work}.  
Second, while exclusion criteria for label noise or limited evidence are straightforward to implement, more detailed annotations for reports with conflicting information and iterating model development to reach minority histology categories will inherently require significant cooperation between the \acs{AI} methods researchers and \acp{SME} who annotate the pathology reports.  

\section*{CONCLUSION}
We have developed a flexible method to characterize sources of error both due to the classification scheme, through our \ac{DAC}, and the identification of individual reports with label noise, hierarchical class complexity, insufficient information, and conflicting evidence.  This method results from combining application of the \textit{\ac{GradInp}} type of local explanations with dimensionality reduction technique like \ac{PCA} to aggregate and extract the most important features for different classes and prediction type. In addition, annotation by top-ranked keywords obtained from the \ac{PCA} analysis provides a visual representation of the global learning patterns which in turn improves interpretability.  We illustrated our method on the automated annotation of major histology classes of lung and breast cancers.  


The explainability method presented in this study has significant potential for applications beyond cancer pathology report classification.  
This method can be adapted to other high-dimensional and complex datasets in healthcare with electronic health records or genomic sequencing results.
Its ability to identify sources of classification errors makes it particularly valuable for improving the reliability and interpretability of \ac{AI}-driven decision-support tools in real-world workflows. 
The abstention mechanism further ensures that high-risk predictions are flagged for manual review, enhancing trust in \ac{AI} systems for critical applications like disease diagnosis/prognosis, and precision medicine. 
This approach could also be extended to non-medical domains with inherent uncertainties, such as legal document analysis, financial risk assessment, and any field requiring interpretable predictions.

\section*{METHODS}

Our goal in developing our global explainability method for application with our \ac{DAC} is to provide practical insights to improve our ability to automatically classify cancer types based on pathology reports.  We expect this method to be useful in other applications and hence characterize each aspect of our interdependent workflow that is necessary to achieve our goal.  After a description of our particular data set, we describe how abstention was built into our \ac{MTCNN} classifier through modification of the loss function, then tuned to achieve 97\% accuracy during training.  We then provide details of our implementation of local explainability through both \textit{\ac{GradInp}} method and \ac{LIME}.  Finally, we provide details of our computation and presentation of the \acs{PCA} of tens of thousands of local explainability results to achieve a global perspective.

\subsection*{Dataset Description}
\label{sec:dataset_description}
The study was done on a corpus of 1,036,784 text cancer pathology reports from the Louisiana and Kentucky Tumor Registries.
Each cancer case (individual tumor) in the dataset has been assigned a ground truth label for each of the tasks. 
\edit{While the classes within individual tasks may have interdependencies and hierarchical relationships, for simplicity, the problem was formulated as independent multi-class classification problem for each task. Each data sample has only one ground truth class for each of the tasks.}
All the reports pertaining to a particular tumor, \edit{that may have multiple reports}, were assigned the same ground truth regardless of the content of the report. 
The maximum allowed length for a report sample was 3000 word tokens; reports were first arranged in the reverse order of words (with the observation that the end of the reports usually contained more important information) and then longer reports were trimmed to 3000 words.

\subsection*{Deep Abstaining Classifier in a Multitask Setting}
\label{sec:abs_classifier}


The \acrfull{DAC} framework, introduced first in Thulasidasan et al.~\cite{thulasidasan2019}, is a framework that allows any \ac{DNN}~\cite{bengio2013DNN, schmidhuber2015DNN, lecun2015DNN} classifier to abstain (or not answer) unless it can statistically guarantee the desired minimum accuracy. This framework, when applied to a \ac{DNN} classifier, appears as a regular classifier, with an extra (abstain) class that behaves as the `none of the above' class but has no ground truth data associated with the class, and a custom loss function that allows abstention during training. The custom loss function allows the \ac{DAC} to learn patterns of confusing or ambiguous samples and abstain on such samples without the need for manual labeling of such examples, while continuing to learn and improve classification performance on the non-abstained samples. 

\subsubsection*{Model architecture}
\begin{figure}[htbp]
	\centering
    \includegraphics[width=0.8\textwidth]{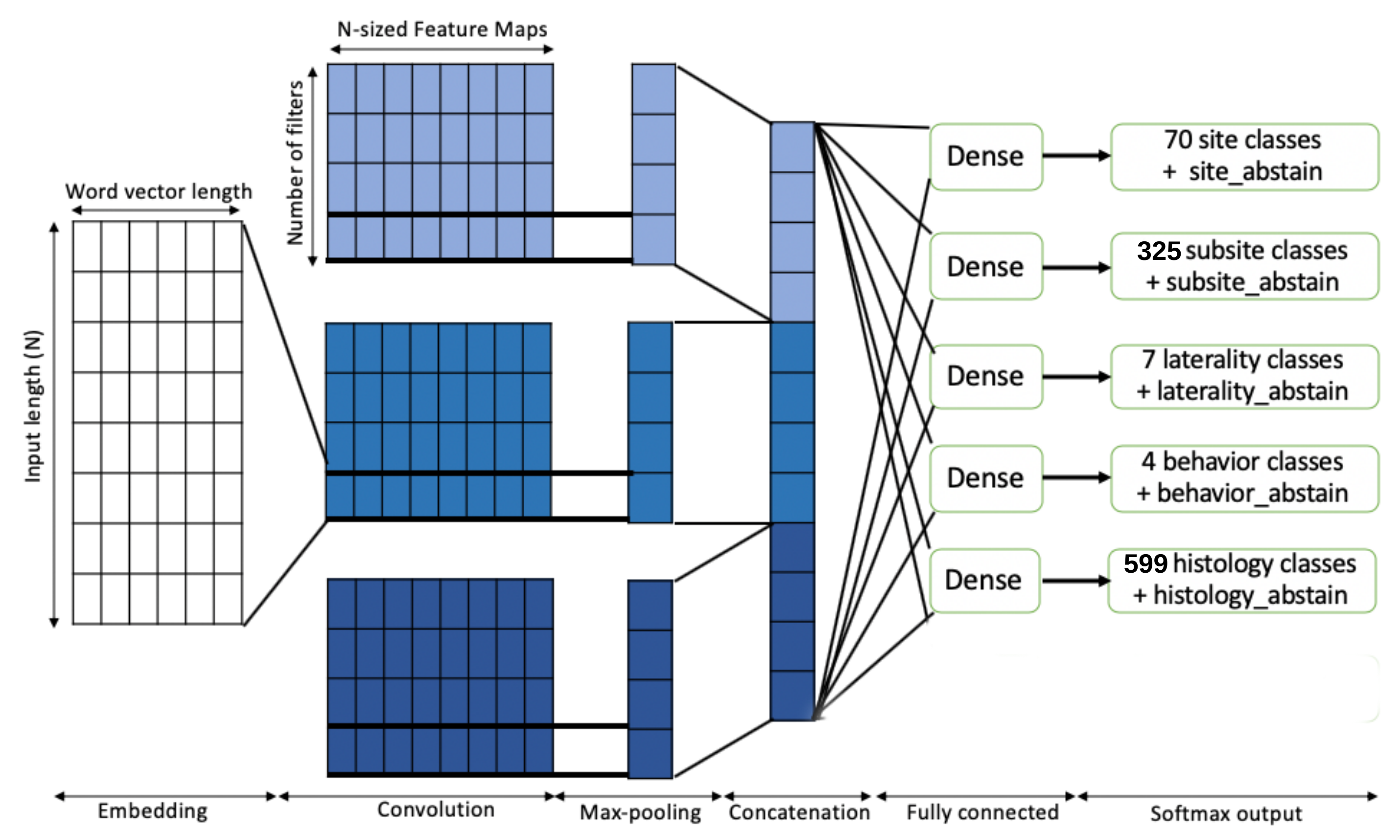}
    \caption{\centering{Our five-task \acrfull{DAC} model architecture includes the input preprocessing, convolutional layers, and prediction layers. Each task includes an additional `abstain’ class, allowing the model to defer decisions when confidence is low.
    }
    }
    \label{fig:model_arch}
\end{figure}

We apply the \ac{DAC} framework to a \acrfull{MTCNN} classifier designed to classify cancer pathology reports. \cref{fig:model_arch} shows the architecture of our \ac{DAC} framework with the \acrfull{MTCNN} base classifier; the output layers for each task have an extra `abstain' class along with all their original classes. The classifier includes three 1-D convolution layers with filter sizes 3, 4, and 5 respectively, 300 filters each, followed by max-pooling, concatenation, and dense (fully connected) layers. This architecture is the same as used in our previous work \cite{dhaubhadel2022} although with a slight variation in the number of tasks, and the number of original classes in individual tasks - primary site (70 classes), subsite (325 classes), laterality (4 classes), behavior (4 classes), and histology type (599 classes), as dictated by the dataset.

\subsubsection*{DAC loss function}
For a given input $x$, denote $y$ to be the predicted class output by the \ac{DNN}. We define $p_i = p_w(y=i\, | \, x)$ (the probability
of the $i\text{th}$ class given $x$) as the $i^\text{th}$ output of the \ac{DNN} that
implements the probability model $p_w(y=i\, | \, x)$ (using a softmax function as its final layer) with $w$ being the set of weight matrices of the \ac{DNN}. For notational brevity, we use $p_i$ in
place of $p_w(y=i\, | \, x)$ when the input context $x$ is clear.

The standard cross-entropy training loss for \acp{DNN} then takes the form
$\mathcal L_\text{standard} = -\sum_{i=1}^k t_i \log p_i$ where $t_i\in\{ 0, 1\}$ 
is the target for the current sample.  The \ac{DAC} has an additional $k+1^\text{st}$ output $p_{k+1}$ which 
denotes
the probability of abstention. We train
the \ac{DAC} with the modified version of the $k$-class
cross-entropy per-sample loss:
$\mathcal{L}(x) =(1-p_{k+1})(-\sum_{i=1}^{k}t_i\log\frac{p_i}{1-p_{k+1}}) + \alpha\log\frac{1}{1-p_{k+1}} \label{abs_loss}$
where k is the number of classes excluding the abstention class, $t_i \in \{0,1\}$ is the true label of training data for class i, $k+1$ is the abstention class, $p_{k+1}$ is the probability of the abstention class and $\alpha$ is the penalty term for abstention.

This loss function behaves like a regular cross-entropy loss on the original classes and adds an additional loss term, scaled by a tuning parameter $\alpha$ that controls the propensity for abstention.   
This parameter is tuned during training to guarantee target accuracy within a specified range while minimizing abstention. 
A very high value of $\alpha$ means a high penalty for abstaining, driving the model towards no abstention. 
Conversely, a very low value of $\alpha$ may drive the model to abstain on everything. 
It is important to note that $\alpha$ is not the same as the abstention rate but a penalty that determines the abstention rate for the data in hand; one can get different abstention rates for the same $\alpha$ value with different subsets of the data.

In a multitask setting, each task has a separate $\alpha$ parameter and the total loss is averaged across all tasks just like a regular multitask classifier.

\subsubsection*{Model training}
\label{sec:exp_setup}
Our \ac{MTCNN} \ac{DAC} model is trained to achieve at least the target accuracy with minimum abstention. 
Earlier versions of the \ac{DAC} were tuned using a combination of accuracy and abstention targets for each task. 
In the latest versions, used for this work, we have implemented training methods that allow targets for either accuracy or abstention alone, in addition to the original mixed targets. 
For this study, we train only for accuracy; for example, in order to guarantee $97\%$ accuracy, we choose accuracy targets of $97.5\% \pm 0.5\%$. 
It is possible to set tighter bounds on the accuracy (or abstention) targets, but this can produce significantly longer training times, especially in a multi-task setting. 
The stopping criteria require that every task satisfy the desired target (which can be set independently for each task), except when a task is able to exceed the target with zero abstention, in which case that task is considered to have satisfied its target. 

Thulasidasan et al.\cite{thulasidasan2019} reported that a \ac{DAC} can learn unlabeled features in the data which may be correlated with label noise. In practice, the label noise is a mix of both uncorrelated ({\itshape e.g.,} labeling inconsistent with the report being classified) and correlated ({\itshape e.g.,} `metastasis' may indicate site labels are unreliable with the sample being confusing and having multiple possible answers) so that perfect empirical identification of misclassified items cannot be achieved.

\subsection*{Local explainability with LIME}
In previous work \cite{dhaubhadel2022} and preliminary analysis for this work, we found that \ac{LIME}~\cite{ribeiro2016lime} technique provided plausible local explanation and a useful visualization, but required extensive sampling (50,000 perturbations for each input sample) and consequently several minutes of computational time per report to provide self-consistent/repeatable results.
We first use the text version of \acrfull{LIME} to generate local (i.e. per sample) explanations and identify which words (in context) were most important (pro or con) in determining the final prediction class for each sample (pathology report). 
\ac{LIME} is provided with a trained \ac{MTCNN} \ac{DAC} model and raw pathology reports, prompting it to return the top 40 words-in-context relevant to identifying the winning class.
It is important to note a hyperparameter that needed significant tuning - the number of perturbations made to the given input sample to gauge the change in output and then estimate the model's sensitivity to change in the input token with a simpler interpretable model (logistic regression). 
The default number of input perturbations - 5000, was insufficient to capture the variation for longer texts such as ours (up to 3000 tokens) and resulted in unstable explanations that changed across multiple runs for the same input sample. 
Upon empirical observation of several \ac{LIME} re-runs for hundreds of samples, 50,000 perturbations appeared to be close to the minimum number of required input perturbations per sample to produce stable\edit{/repeatable} explanation. 

However, this increase in number of perturbation from 5,000 to 50,000 significantly increased the computation time to generate \ac{LIME} explanation from a few seconds to a few minutes per sample depending on the input length.

\subsection*{Local explainability with \textit{\acrfull{GradInp}} explainability technique}
Gradient-based local explainability techniques identify which parts of the input most influenced the model’s decision. For example, saliency maps~\cite{simonyan2014deepinside}, originally proposed for images to generate pixel wise `importance score', have been applied \edit{to} \ac{NLP} by applying the partial derivative of the output with respect to the word embedding of the input words such that the gradients represent the sensitivity of the word across each embedding dimension. The gradients across the embedding dimension can be aggregated in many ways (averaged, summed, etc.) to obtain the overall sensitivity of a particular word ~\cite{denil2014, li2016saliencyNLP}.

We generate our local explanations by applying a slight variation of the \textit{\ac{GradInp}} method \cite{denil2014} by taking the dot product of the re-centered input embedding (i.e. word embedding - embedding centroid) with the gradients (partial derivative) of the embeddings. 
The re-centering is required because the default coordinate system of the embedding space is centered on the first word in the dictionary, so we use the embedding centriod as the origin. 
The use of the \textit{\ac{GradInp}} provides an analogy with the local explainability methodology \ac{LIME}~\cite{ribeiro2016lime} 
, where individual words in context are selectively masked/removed randomly (not bag of words) to test the sensitivity of the model prediction to these masked words; a strongly aligned positive gradient is equivalent to addition of the keyword and negative gradient is equivalent to the subtraction of the keyword. 
This method provides different `importance scores' to different instances of the same features (words) depending on the context in which they are used as the gradient varies a lot depending on the context.
This projection effectively removes gradients which are orthogonal to the word embedding; but given the dimensionality of the embedding space and the size of the dictionary, there is at present no simple way to interpret these gradients, which are effectively rotating the word embedding toward some unknown location. 
\edit{The \textit{\acs{GradInp}} technique
could be computed in 1-2 seconds per sample, i.e. 0.83 hours (50 minutes) to generate 1000 local explanations in contrast to \ac{LIME} that would take roughly 80 hours of compute time, with meaningful local explanations that compared well to \ac{LIME} (see Figure \ref{fig:lime_vs_saliency}). The significantly smaller compute time with the \textit{\acs{GradInp}} technique  allows aggregation of hundreds of thousands of local explanations within a reasonable amount of time. 
Hence, for the rest of the work, we used \textit{\acs{GradInp}} technique instead of \ac{LIME} for generating local explanations.}

\subsection*{Global explainability with \texorpdfstring{\acs{PCA}}{PCA} applied to local explanations}
While individual local explanations can be extremely useful in understanding the model's reasoning for a particular classification decision, it becomes impractical to examine local explanations once the data size grows to thousands (millions in our case).  Hence, in order to obtain meaningful global insights from thousands of local explanations for a given predicted class, group of classes, or prediction type (correct classification, confusion, etc.), we build a global explainability method that enables systematic disentangling of local explanations to identify global patterns learned by the model. This method involves constructing \editone{an \acrfull{ALE} matrix} using the local explanations, and applying dimensionality reduction techniques like \ac{PCA} to the \editone{\ac{ALE} matrix} to identify the most significant features for a global concept of interest. This global concept may be a class or a combination of classes (ground truth and/or predicted). In this work, we demonstrate our global explainability method on local explanations obtained from the \textit{\ac{GradInp}} technique described above.

For our study, we start by picking all the reports correctly classified as lung and breast cancers for the site task (to avoid confusion due to site ambiguity).
We then separate these reports into groups of abstained and non-abstained for the histology classification task. 
We then sub-group the abstained and non-abstained reports based on all possible combinations of their ground truth and predicted classes to understand and distinguish the types of confusion in each group. 
The predicted classes are identified as the classes with the highest softmax scores for non-abstained reports. For abstained samples, however, the `predicted classes' are classes with the second highest softmax scores, i.e. the potential predictions in the absence of abstention, are used.
We only consider the four largest histology classes for each of the cancer sites for visual tractability of our results and also because the top four classes accounted for over 80\% samples for a given site.  
Finally, we cap each category of ground truth - prediction combination for each of the non-abstained and abstained groups to 1000 reports, ensuring the best possible representation of minority ground truth - prediction combination.

After deciding which local explanation samples to consider for global explainability, the first step in our global explainability method involves constructing \editone{the \ac{ALE} matrix} such that the rows represent individual data samples (i.e. reports) and the columns represent all features in the feature space (i.e. all words in the data dictionary). 
The value in each position $(i,j)$ of this local explanation matrix corresponds to the explainability weight or `importance score' of word $j$ in report $i$.
Since different instances of the same feature (word) in a given sample (report) may have different explainability weights depending on the context in which the feature is used, all the weights for the given feature within a sample are summed to compute the overall contribution of the feature towards the classification decision for the sample. 
In principle, one could sum the absolute values of the individual weights to avoid cancellation of strong positive and negative occurrences of the same feature, but in practice the weights of a feature within a single sample are generally of the same sign.

This \editone{\ac{ALE} matrix} is very sparse, with 0's for most irrelevant features, particularly when examined in a context of a specific class or misclassification. To reduce sparsity and remove seemingly irrelevant features, the matrix is then truncated to only include a subset of the features that have \emph{`reasonable importance'} across several samples. To do this, we only include features \edit{(i.e. words)} for which the sum of absolute values of explainability weights across all samples under consideration are above a certain \edit{explainability weight} threshold, chosen to yield tractable matrix sizes while still capturing relevant (non-sparse) features. 

\Ac{PCA}~\cite{pearson1901pca, jolliffe2016pca} is a widely used dimensionality reduction technique to reduce data with large dimensions into smaller dimensions while retaining the maximum possible information (maximize explained variance) about the dataset. We apply \ac{PCA} to the truncated local explainability matrix to extract features that capture the majority of local explainability space across thousands of samples along the direction of the first two or three principal components. We then \editone{analyze and} visualize the \ac{PCA} results along the top principal component axes\editone{, accounting for at least 90\% of total variance cumulatively,} to highlight patterns across the samples (reports) based on their \emph{`most important'} features reduced to the principal components. 
\edit{The visualizations are done with color coding based on two approaches. First, on the basis of ground truth and predicted classes, which highlights types of confusion between distinct classes, their interdependencies, as well as label noise. Second, based on the most `important' word level features extracted from the local explainability space with our the \ac{PCA} analysis i.e. the largest contributors to the eigenvectors as shown in Tables \ref{tab:lu_ev_table} and \ref{tab:br_ev_table} respectively. This visualization highlights the reasons for confusion including the presence of conflicting/ambiguous information on multiple classes, and the hierarchical relationships between some classes.}
To illustrate the effectiveness of our method, we demonstrate such global patterns for correct/incorrect classifications as well as abstentions in several different classes of histology for lung and breast cancers.

\edit{The main steps of our global explainability method are summarized below:
 \begin{itemize}[noitemsep, nolistsep]
    \item Construct an \editone{\ac{ALE} matrix} such that the rows represent individual reports, and the columns represent the available feature space i.e. all words in the data dictionary that made it to the top N most important words at least one of the local explanations. 
    \begin{itemize}[noitemsep, nolistsep]
        \item The value $w$ in position $(i,j)$ of the local explanations matrix corresponds to the sum of local explainability weights for all occurrences $k$ of word $j$ in report $i$: $w_{i,j} = \sum_k w_{i,j,k}$.
    \end{itemize}
    \item Truncate the matrix to only include a subset of the words that have weights with a sum of in absolute value across all reports $i$ above a certain minimum value $T$ that is determined empirically: $\sum_i |w_{i,j}| > T$; $T$ is chosen to ensure coverage of relevant words while retaining computational tractability.
    \item Apply PCA to this subsetted matrix and visualize the first two principal components such that each point in the 2D plot represents a single report along the given principal components
    \begin{itemize}[noitemsep, nolistsep]
        \item Color code the individual points, each representing an individual report, based on different criteria including true and predicted labels, presence of certain deterministic keywords.
    \end{itemize}
\end{itemize}
}


\section*{Acknowledgements}
We would like to acknowledge James Mac Hyman for his valuable feedback on the manuscript.

This work has been supported in part by the \ac{JDACS4C} program established by the \acrfull{DOE} and the \acrfull{NCI} of the \ac{NIH}. This work was performed under the auspices of the \ac{DOE} by Argonne National Laboratory under Contract DE-AC02-06-CH11357, Lawrence Livermore National Laboratory under Contract DEAC52-07NA27344, Los Alamos National Laboratory under Contract DE-AC5206NA25396, and Oak Ridge National Laboratory under Contract DE-AC05-00OR22725.

The authors would like to acknowledge the contribution to this study from other staff in the participating central cancer registries. These registries are supported by the  \acrfull{NCI}’s \acrfull{SEER} Program, the Centers for Disease Control and Prevention’s National Program of Cancer Registries (NPCR), and/or state agencies, universities, and cancer centers.  

The participating central cancer registries include the following: 
\begin{itemize}[noitemsep, nolistsep]
    \item Kentucky working under contract numbers SEER: HHSN261201800013I/HHSN26100001 and NPCR: NU58D007144 
    \item Louisiana working under contract numbers SEER: HHSN261201800007I/HHSN26100002 and NPCR: NU58DP0063. 
\end{itemize}

\section*{Authors' contributions}
SD, JMY, and BHM did the bulk of the coding and writing of the manuscript. 
KG served as the \acrfull{SME} for the study.
AS and JPG provided assistance with implementation and running the pipeline.
XCW and EBD were responsible for providing the data.
SD,
JMY,
TE, HAH, BHM, and TB contributed to the study design and provided guidance and ideas related to the analyses throughout the project.

\section*{Competing interests}
The authors declare no competing interests.

\section*{Data availability}
The study was conducted on sensitive \ac{PHI} from the \ac{NCI} and cannot be made available without appropriate data usage agreement and \ac{IRB} approval.

\section*{Ethics declarations}
None

\newpage
\bibliography{refs}


\printnoidxglossary

\end{document}


\onecolumn

\beginsupplement

\section*{Confusion matrices}

\begin{figure}[htb]
	\centering
        \caption{Confusion matrices showing the 20 largest histology classes of lung cancer \textbf{(top)} and breast cancer \textbf{(bottom)}. 
        The \textbf{left} panel shows the results of a baseline \acs{MTCNN} classifier while the \textbf{right} panel shows the results of our \acs{MTCNN} \acs{DAC} classifier. 
        The figures are color coded according to the counts in them on a logarithmic scale: blacks and darker blues indicating higher counts. 
        }
        \label{fig:confusion_matrix}
        
	\includegraphics[width=0.45\textwidth,trim={0 0 0 0cm},clip]{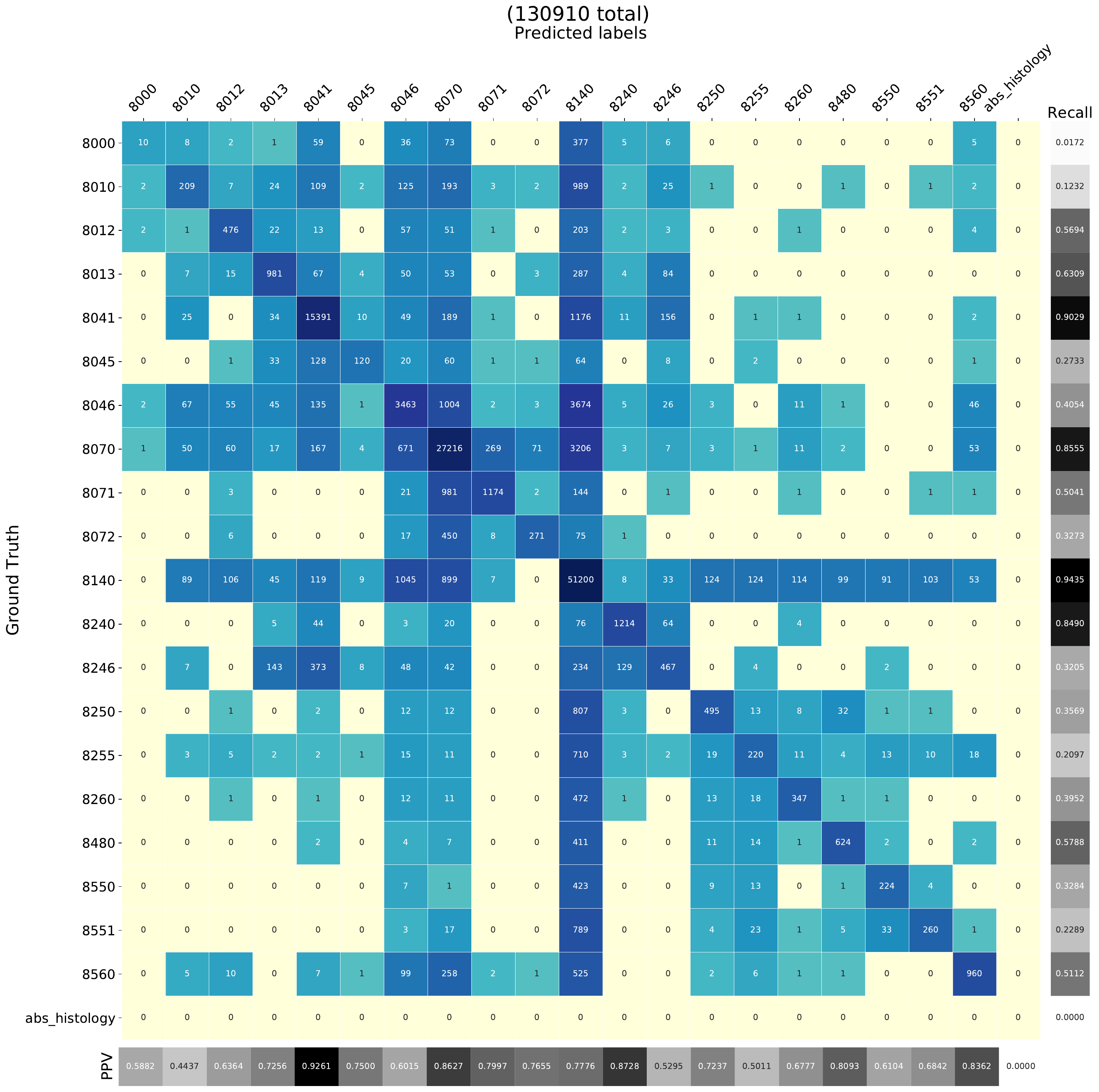}
 	\includegraphics[width=0.45\textwidth,trim={0 0 0 0cm},clip]{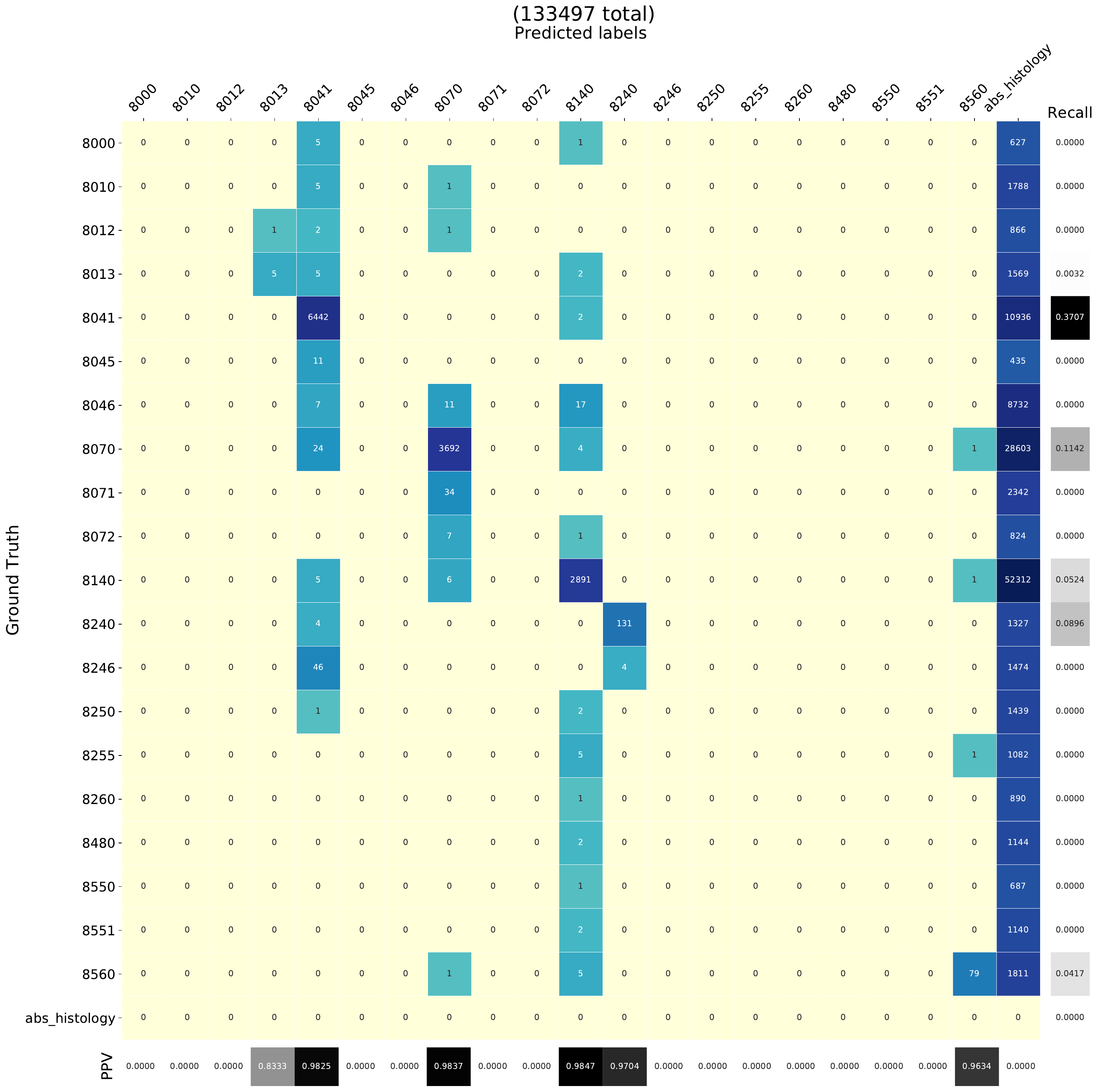}
        
        \bigskip
	\includegraphics[width=0.45\textwidth,trim={0 0 0 0cm},clip]{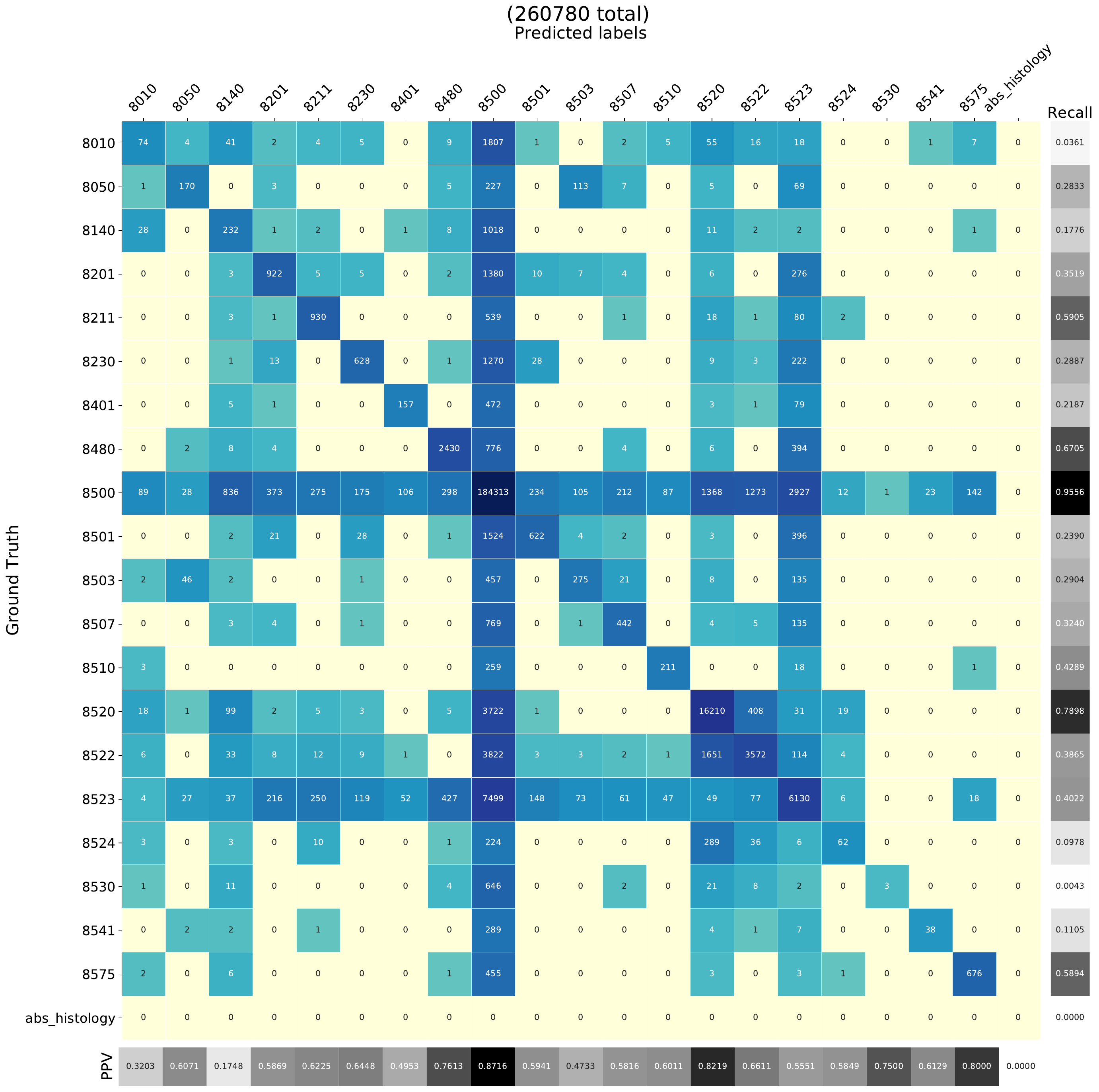}
 	\includegraphics[width=0.45\textwidth,trim={0 0 0 0cm},clip]{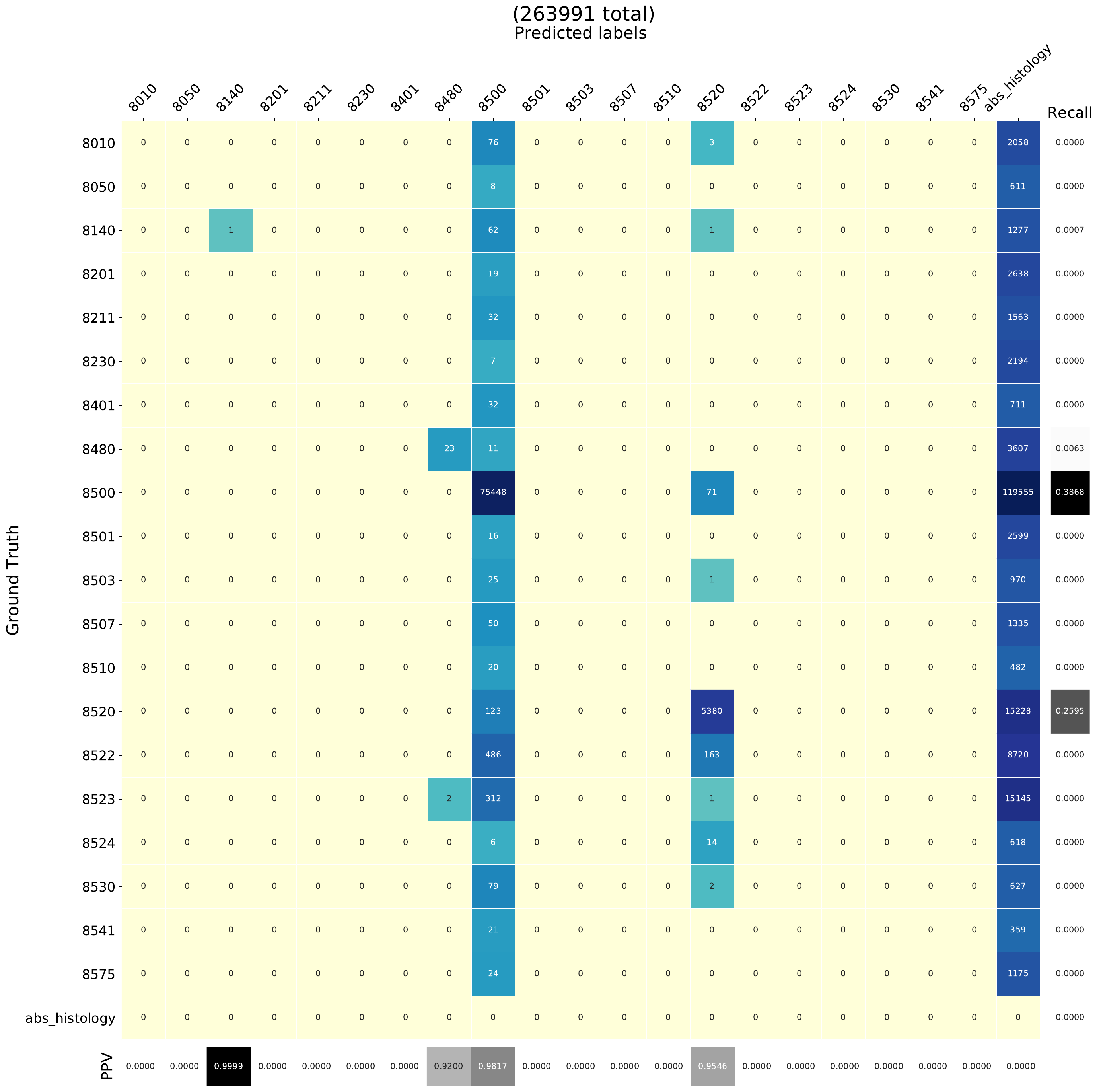}
\end{figure}

\newpage
\section*{\Acrfull{GradInp} vs \ac{LIME}}
\begin{figure}[htbp]
	\centering
        \caption{Comparison of \ac{GradInp} explanations (y-axis) with \ac{LIME} explanations(x-axis) \textbf{(left)} lung cancer adenocarcinoma and \textbf{(right)} breast cancer ductal carcinoma. 
        Given \ac{GradInp} and \ac{LIME} explanations for the same set of reports, the axes show the number of reports with report-level positive or negative weights for the given word. 
        Black dots show explanations for non-abstained reports and red triangles show explanations for abstained ones
        }
        \label{fig:lime_vs_saliency}
        
	\includegraphics[width=0.49\textwidth, trim={1.5cm 2cm 0 0},clip]{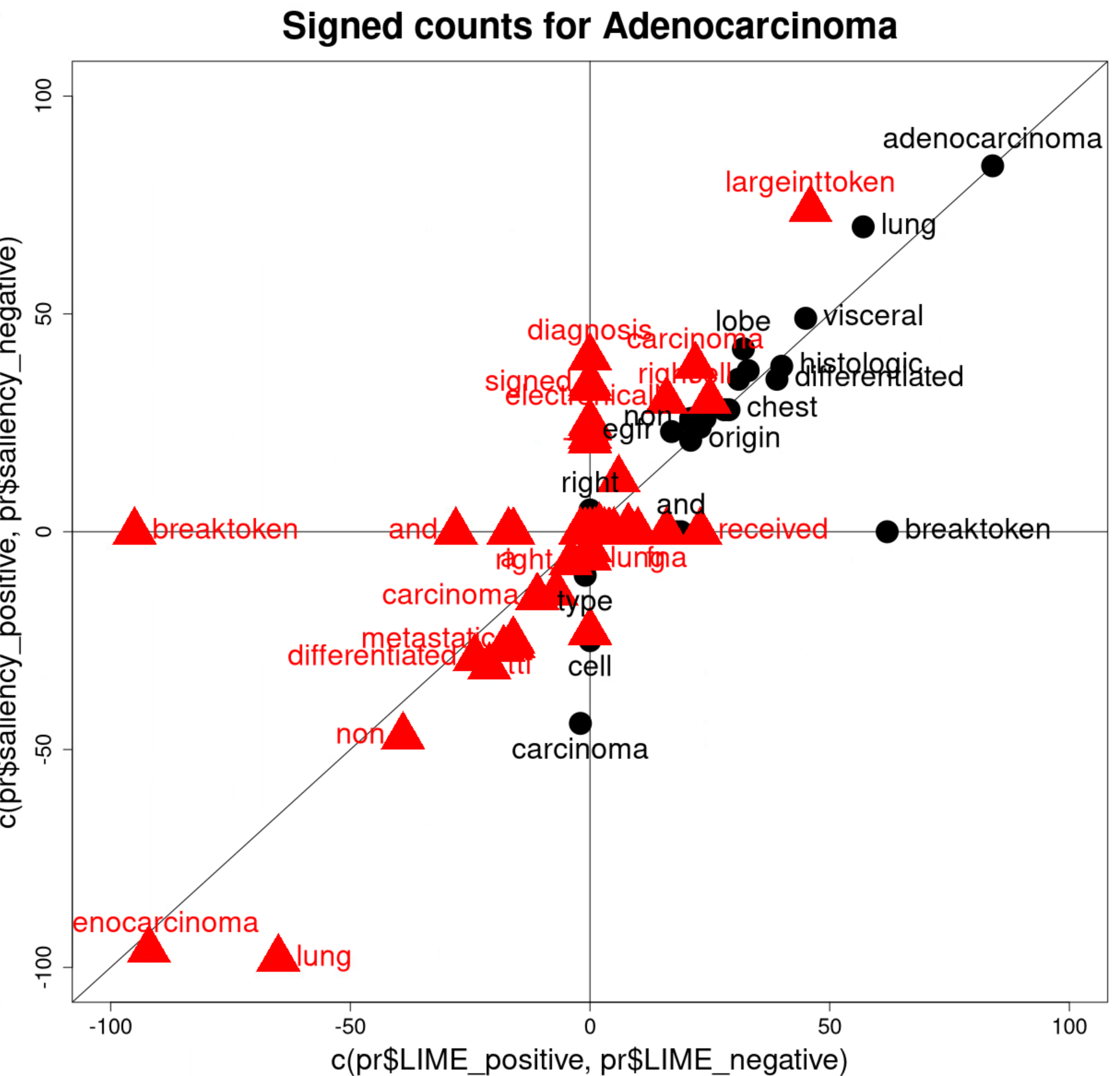}
 	\includegraphics[width=0.49\textwidth, trim={1.5cm 2cm 0 0},clip]{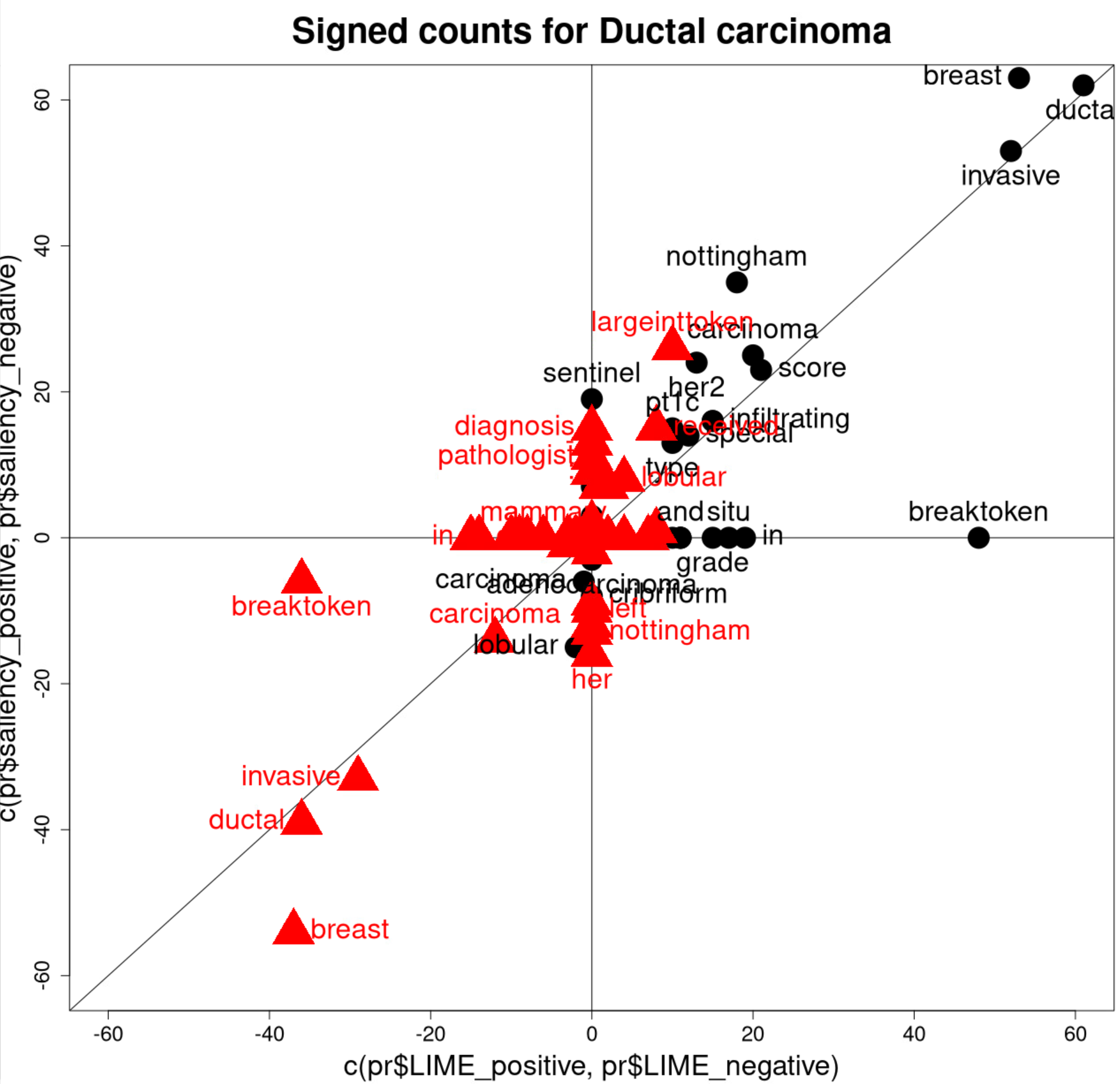}
        
\end{figure}

\section*{Distribution of local explanations}
\begin{figure}[htbp]
    \centering
        \begin{subfigure}{0.99\textwidth}
        \includegraphics[width=0.99\columnwidth]{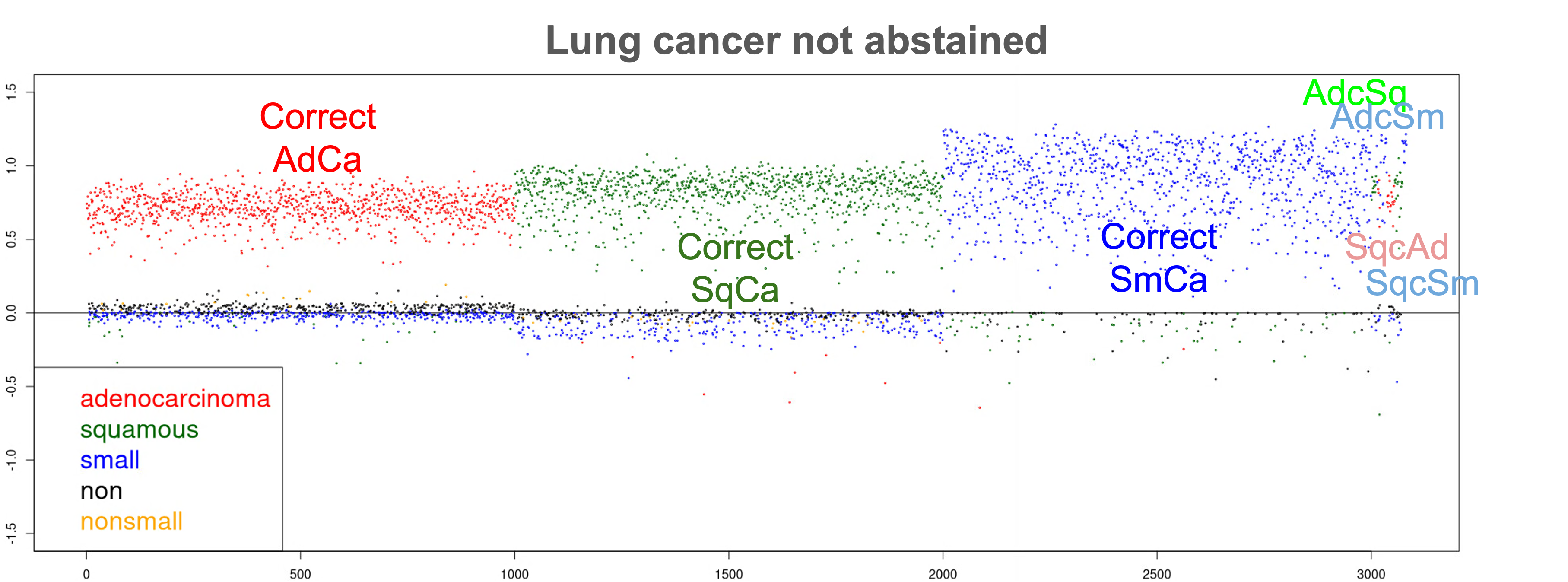}
		\vspace*{-1\baselineskip}
            \caption{}
            \label{fig:lu_nabs_dist}
	\end{subfigure}
        \begin{subfigure}{0.99\textwidth}
        \includegraphics[width=0.99\columnwidth]{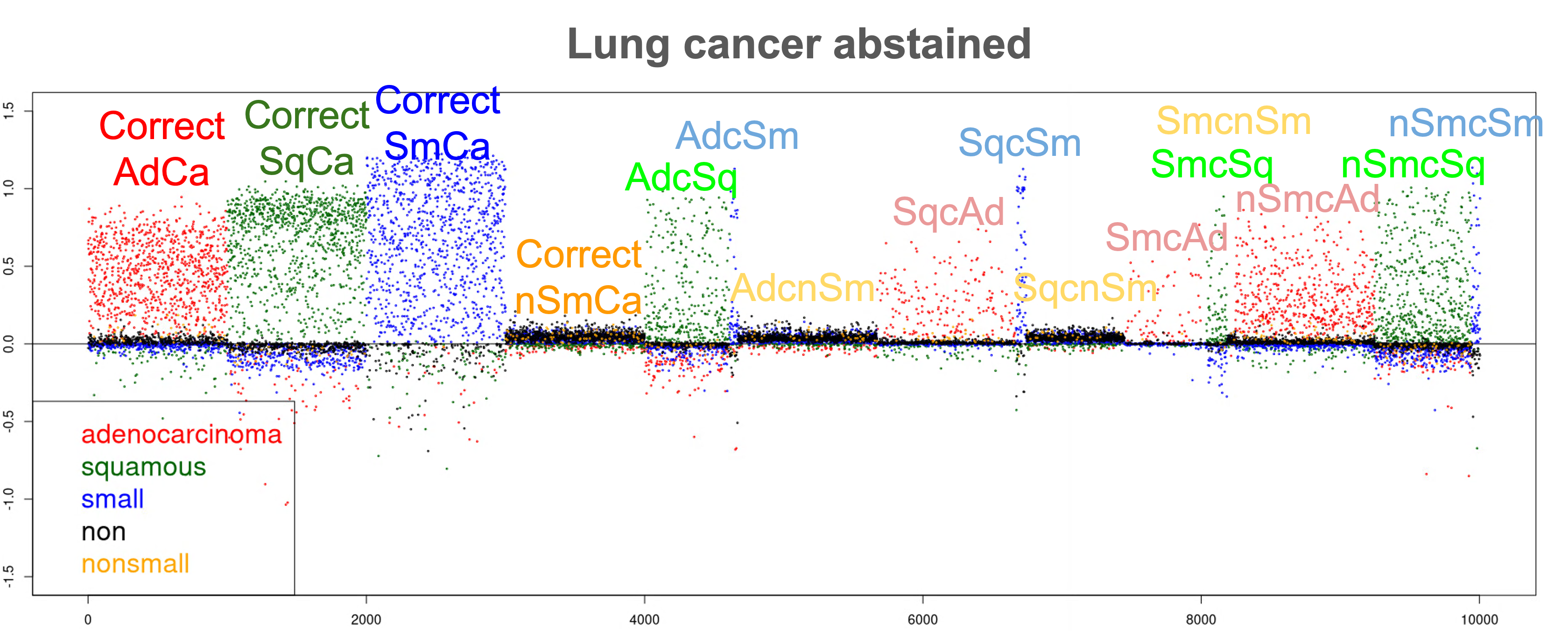}
            \vspace*{-1\baselineskip}
            \caption{}
		\label{fig:lu_abs_dist}
	\end{subfigure}
        
    \caption{Local explainability weights distribution in non-abstained (\textbf{top}) and abstained (\textbf{bottom}) reports for four largest histology classes of \textbf{lung cancer} reports. Correct classifications are shown as "Correct $<$class$>$" while incorrect classifications are coded as $<ground\_truth>c<predicted>$.
    }
    \label{fig:lu_dist}
\end{figure}

\begin{figure}[htbp]
    \centering
        \begin{subfigure}{0.99\textwidth}
        \includegraphics[width=0.99\columnwidth]{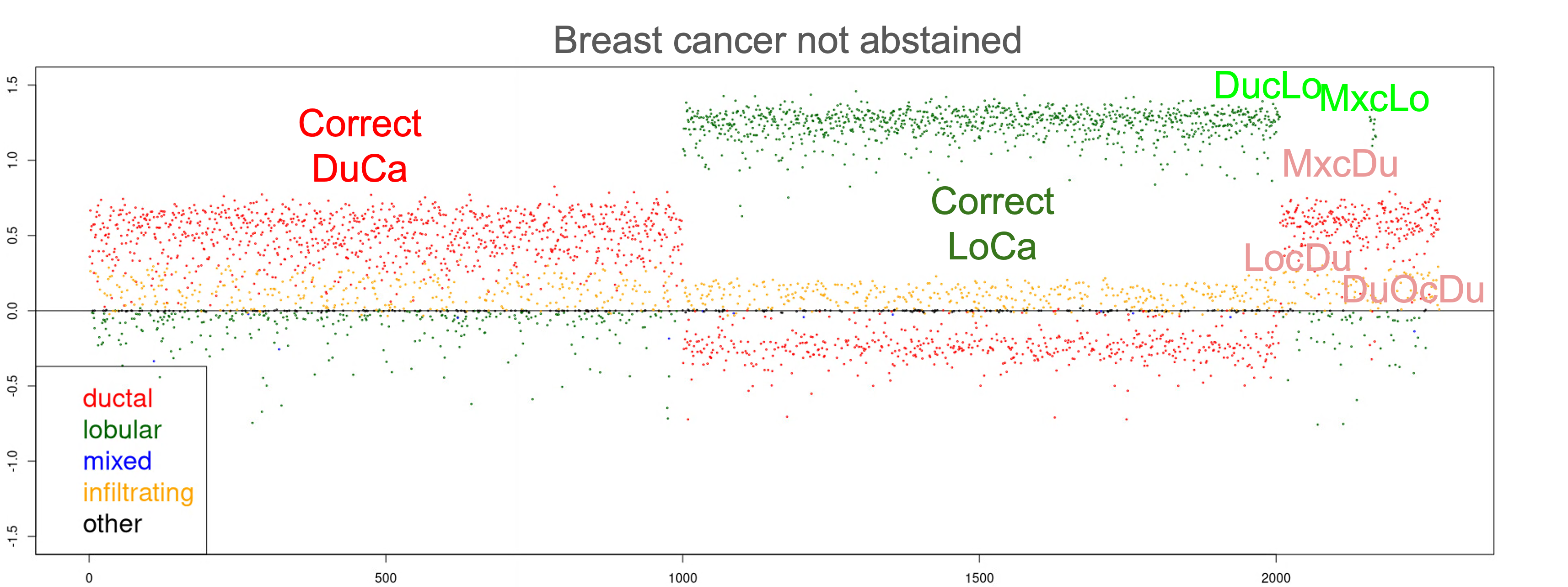}
		\vspace*{-1\baselineskip}
            \caption{}
            \label{fig:br_nabs_dist}
	\end{subfigure}
        \begin{subfigure}{0.99\textwidth}
        \includegraphics[width=0.99\columnwidth]{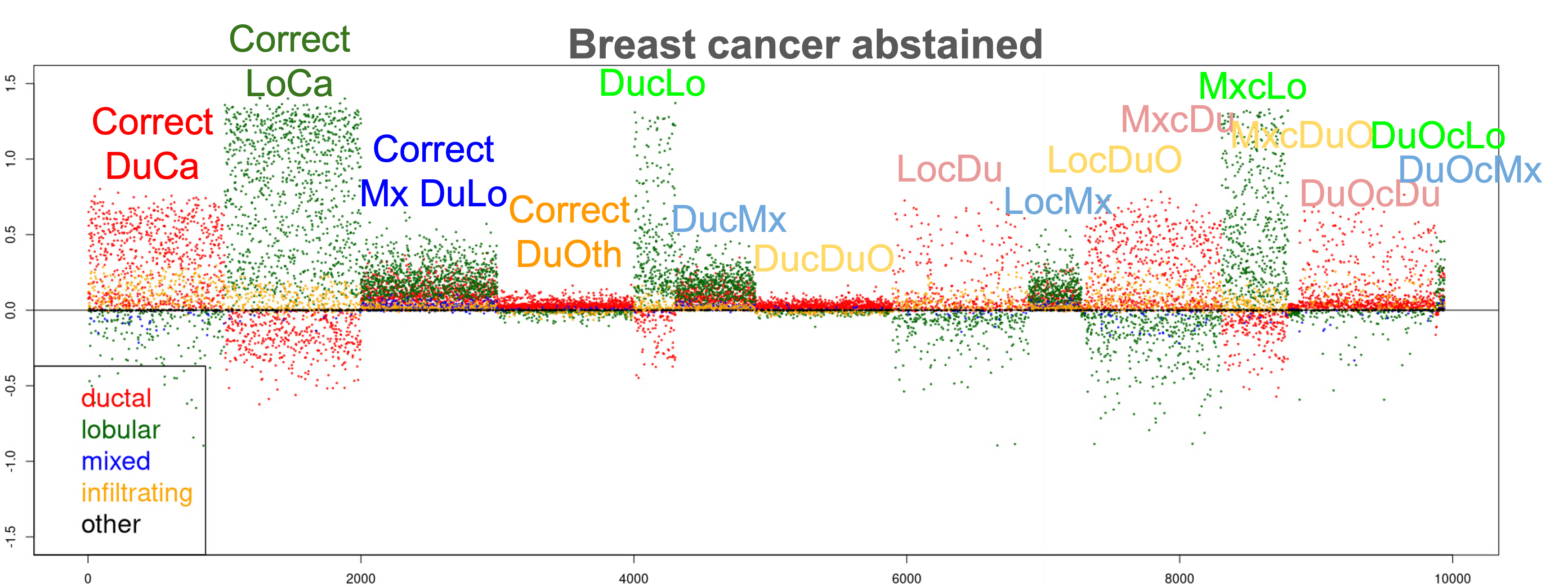}
            \vspace*{-1\baselineskip}
            \caption{}
		\label{fig:br_abs_dist}
	\end{subfigure}
        
    \caption{Local explainability weights distribution in non-abstained (\textbf{top}) and abstained (\textbf{bottom}) for four largest histology classes of \textbf{breast cancer} reports. Correct classifications are shown as "Correct $<$class$>$" while incorrect classifications are coded as $<ground\_truth>c<predicted>$.
    }
    \label{fig:br_dist}
\end{figure}

Fig. \ref{fig:lu_dist} and Fig. \ref{fig:br_dist} show importance scores for the most-important words across $\sim$13,000 reports representing the correct and incorrect classifications in the four largest histology classes of lung and breast cancer respectively. 
    We plot such distributions for both non-abstained as well as abstained samples. The importance scores are computed with respect to the highest softmax scores for the non-abstained samples. For the abstained ones, the importance scores are computed with respect to the second highest softmax scores i.e. the model's preferred choice in the absence of abstention. 
    Comparison of the distribution of the non-abstained (top) and abstained (bottom) samples shows much higher level of uncertainty in the abstained samples. Important words for correctly classified non-abstained samples appear to be in uniform distribution and deterministic with relatively large positive importance scores for the class-specific words and sometimes corresponding negative scores for words associated with a different class. 
    The distribution of the important words for incorrectly classified non-abstained samples shows more interesting trends, with a high positive score for words associated with the predicted class and sometimes a negative score associated with the ground truth class, suggesting that these samples either have information only on the predicted class or conflicting information between the predicted and ground truth class.
    Plots of the importance score distribution for the second choice in the abstained samples (lower panel) look much noisier, although following similar pattern as the non-abstained ones. The potential correct classifications have lower positive weights for the class-specific important words and often corresponding negative weights for keywords belonging to different classes, suggesting higher uncertainty. For abstained samples, the large overlap in the distribution between the potentially correct and incorrect classifications, also supports the model's reasoning for abstaining on those samples.

\newpage

\begin{table}[h]
    \centering
    \caption{Eigenvector matrix for the top 10 words based on magnitude along PC1 and PC2 for \textbf{lung cancer}; variance explained by each principal component are provided in parentheses; eigenvalues for the first 5 principal components: 0.1249, 0.0960, 0.0484, 0.0059, 0.0032}
    \label{tab:lu_ev_table}
    
    \begin{tabular}{|c|c|c|c|c|c|} 
    \hline
           &  PC1 (39.14\%) &  PC2 (30.08\%) &  PC3 (15.17\%) &  PC4 (1.85\%) &  PC5 (1.01\%) \\
    \hline
        small  &  -0.8625  &  -0.2034  &  0.2566  &  -0.0252  &  -0.0476  \\
        squamous  &  0.3971  &  -0.7892  &  0.3861  &  -0.0620  &  -0.0071  \\
        adenocarcinoma  &  0.1312  &  0.4882  &  0.8057  &  -0.0540  &  -0.1543  \\
        carcinoma  &  -0.1509  &  -0.2284  &  0.0826  &  -0.0182  &  0.0930  \\
        cell   &  -0.2083  &  -0.1566  &  0.0825  &  -0.0101  &  -0.0026  \\
        pulmonary  &  0.0137  &  0.0733  &  0.1348  &  -0.6205  &  0.5283  \\
        bronchial  &  -0.0123  &  -0.0527  &  0.0389  &  -0.1566  &  -0.5748  \\
        consistent  &  -0.0523  &  -0.0110  &  0.0445  &  -0.0087  &  -0.0271  \\
        differentiated  &  0.0350  &  -0.0277  &  0.0345  &  0.0203  &  -0.0242  \\
        metastatic  &  -0.0440  &  0.0063  &  0.0523  &  -0.0138  &  0.4156  \\
    \hline
    \end{tabular}
\end{table}

\begin{table}[h]
    \centering
    \caption{Eigenvector matrix for the top 10 words based on magnitude along PC1 and PC2 for \textbf{breast cancer}; the variance explained by each principal component are provided in parentheses; eigenvalues for the first 5 principal components: 0.2002, 0.0478, 0.0053, 0.0035, 0.0017}
    \label{tab:br_ev_table}
    
    \begin{tabular}{|c|c|c|c|c|c|}
    \hline
           &  PC1 (69.75\%)  &  PC2 (16.65\%)  &  PC3 (1.86\%)  &  PC4 (1.23\%) &  PC5 (0.58\%) \\
    \hline
        lobular & 0.9554 & 0.1829 & 0.1885 & -0.02786 & -0.0323 \\
        ductal & -0.2665 & 0.8170 & 0.4647 & -0.07489 & -0.0382 \\
        breast & 0.0443 & 0.4455 & -0.7092 & -0.1118 & 0.1228 \\
        invasive & 0.0235 & 0.2174 & -0.1569 & 0.6936 & -0.0452 \\
        carcinoma & 0.0485 & 0.0713 & -0.0645 & -0.0290 & -0.0654 \\
        left & 0.0512 & 0.0571 & -0.1630 & -0.0055 & 0.3937 \\
        infiltrating & 0.0097 & 0.0720 & -0.1087  &  -0.6794 & -0.0282 \\
        score  &  0.0088  &  0.0628  &  -0.0576  &  0.0716  &  -0.0570  \\
        :  &  -0.0075  &  -0.0601  &  0.0195  &  -0.0245  &  0.0844  \\
        nottingham  &  0.0050  &  0.0567  &  -0.0576  &  0.0960  &  -0.0594  \\
        right & 0.0276 & 0.0467 & -0.1179 & -0.0039 & -0.0554 \\
        features & 0.0026 & -0.0155 & 0.1821 & 0.0053 & -0.3851 \\
        mixed & 0.0012 & -0.0006 & 0.0265 & 0.0079 & -0.0426 \\
    \hline
    \end{tabular}
\end{table}

\clearpage

\printnoidxglossary
